\documentclass{article}

\usepackage[nonatbib,preprint]{neurips}

\usepackage[square,sort,comma,numbers]{natbib}

\usepackage[utf8]{inputenc} 
\usepackage[T1]{fontenc}    
\usepackage{hyperref}       
\usepackage{url}            
\usepackage{booktabs} 
\usepackage{amsfonts}       
\usepackage{nicefrac}       
\usepackage{microtype}      
\usepackage{xcolor}         
\usepackage{amsmath}
\usepackage{multirow}
\usepackage{mathtools}
\usepackage{pdfpages}
\usepackage{CJKutf8}
\usepackage{graphicx}
\usepackage{rotating}
\usepackage{tablefootnote}

\usepackage[final]{changes}

\usepackage{float}
\floatstyle{plaintop}
\restylefloat{table}

\usepackage{colortbl}
\definecolor{lightgray}{gray}{0.9}

\newcommand{\ja}[1]{%
  \begin{CJK}{UTF8}{ipxm}#1\end{CJK}%
}

\title{Evolutionary Optimization of Model Merging Recipes}

\author{%
  Takuya Akiba, Makoto Shing, Yujin Tang, Qi Sun, David Ha\\
  Sakana AI\\
  Tokyo, Japan\\
  \texttt{\{takiba,mkshing,yujintang,qisun,hadavid\}@sakana.ai} \\
}

\begin{document}
\maketitle
\vskip -0.195in 
\begin{abstract}
\vskip -0.05in 
\added{Large language models (LLMs) have become increasingly capable, but their development often requires substantial computational resources.}
\deleted{We present a novel application of evolutionary algorithms to automate the creation of powerful foundation models.} While model merging has emerged as a \added{cost-effective} promising approach for \deleted{LLM development}creating new models by combining existing ones\deleted{, due to its cost-effectiveness}, it currently relies on human intuition and domain knowledge, limiting its potential. Here, we propose an evolutionary approach that overcomes this limitation by automatically discovering effective combinations of diverse open-source models, harnessing their collective intelligence without requiring extensive additional training data or compute. Our approach operates in both parameter space and data flow space, allowing for optimization beyond just the weights of the individual models. This approach even facilitates cross-domain merging, generating models like a Japanese LLM with Math reasoning capabilities. Surprisingly, our Japanese Math LLM achieved state-of-the-art performance on a variety of established Japanese LLM benchmarks, even surpassing models with significantly more parameters, despite not being explicitly trained for such tasks. Furthermore, a culturally-aware Japanese VLM generated through our approach demonstrates its effectiveness in describing Japanese culture-specific content, outperforming previous Japanese VLMs. This work not only contributes new state-of-the-art models back to the open-source community, but also introduces a new paradigm for automated model composition, paving the way for exploring alternative, efficient approaches to foundation model development.
\end{abstract}

\section{Introduction}
\vskip -0.05in 

Model merging~\cite{goddard2024merge,labonne2024merge}, a recent development in the large language model (LLM) community, presents a novel paradigm shift. By strategically combining multiple LLMs into a single architecture, this exciting development has captured the attention of researchers due to its key advantage: it requires no additional training, making it an incredibly cost-effective approach for developing new models. This accessibility has fueled a surge in interest and experimentation with model merging. The Open LLM Leaderboard~\cite{openllmleaderboard} is now dominated by merged models, showcasing its potential for democratizing foundation model development.

However, model merging is considered by many to be a form of black art or alchemy, relying on the model maker’s intuition and instincts about model selection and merging recipes to create and refine a new model that performs well for a particular task. Furthermore, the model maker is often required to have some domain knowledge for the various different benchmark tasks.
Given the large diversity of open models and benchmarks in the community, human intuition can only go so far, and we believe a more systematic approach for discovering new model combinations will take things much further.

We believe evolutionary algorithms will be able to discover more effective model merging solutions, and thus provide a path for automating the creation of more capable models.
As a step towards this direction, in this work, we show that evolution can be employed to discover novel and unintuitive ways to merge \replaced{multiple}{various} models to produce new models with a new combined ability.
In this work, we present a methodology that leverages evolutionary algorithms to facilitate the merging of foundation models. Our approach is distinguished by its ability to navigate both parameter space (weights) and the data flow space (inference path), proposing a framework that integrates these two dimensions.

This work makes several key contributions to the field of foundation model development:

\begin{enumerate}
\item \textbf{Automated Model Composition}: We introduce \textbf{Evolutionary Model Merge}, a general evolutionary method to automatically discover \replaced{effective}{optimal} combinations of \replaced{selected}{diverse open-source} models for creating new foundation models with user-specified capabilities. This approach harnesses the collective intelligence of existing open models, enabling the creation of powerful models without the need for extensive training data or compute.
\item \textbf{Cross-Domain Merging}: We demonstrate that our method can discover novel ways to merge models from disparate domains (e.g., non-English language and Math, non-English language and Vision), potentially exceeding the capabilities achievable through conventional human design strategies.
\item \textbf{State-of-the-Art Performance}: We showcase the effectiveness of our method by automatically generating a Japanese LLM with Math reasoning capability and a Japanese Vision-Language Model (VLM). Notably, both models achieve state-of-the-art performance on various benchmarks, even without explicit optimization for those tasks.
\item \textbf{High Efficiency and Surprising Generalizability}: We observe that our 7B parameter LLM surpasses the performance of some previous 70B parameter Japanese LLMs on benchmark datasets, highlighting the high efficiency and surprising generalization capability of our approach. We believe this model can serve as a strong general-purpose Japanese LLM.
\item \textbf{Culturally-Aware VLM}: The generated Japanese VLM achieves top results when tested on a domestically-sourced dataset of Japanese image-description pairs, demonstrating its ability to handle Japanese culture-specific content.
\end{enumerate}

We are committed to open science and are excited to open-sourc\added{e}\deleted{ing} our \texttt{EvoLLM-JP} and \texttt{EvoVLM-JP}, two state-of-the-art Japanese foundation models, to the community, enabling further research and development in the field. Our work challenges the conventional paradigm of expensive model development by demonstrating that our evolutionary-based method can produce competitive models without relying on gradient-based training. This paves the way for exploring alternative, potentially more efficient, approaches to foundation model development.
 
\subsection{Background and Related Work}
\vskip -0.05in 
\subsubsection{Overview of Model Merging.}
\vskip -0.05in 
Model merging offers a novel approach to leverage the strengths of multiple pre-trained models. It allows us to combine task-specific models, each potentially fine-tuned for a particular downstream task, into a single unified model. This technique stands in contrast to traditional transfer learning, where a pre-trained model is further fine-tuned for a new task. While transfer learning offers advantages like improved performance and faster convergence, the resulting models are typically limited to single tasks. Model merging, on the other hand, strives to create a versatile and comprehensive model by combining the knowledge from multiple pre-trained models, potentially yielding a model capable of handling various tasks simultaneously.

While fine-tuning an existing model involves training an existing model further on a specialized dataset, model merging methods work by combining existing models without any additional gradient-based training, making it very cost-effective in terms of compute requirements. Many in the machine learning community, including researchers, hackers, and hobbyists alike, are using both fine-tuning and model-merging to create the best models for their purposes. In fact, this collaborative way of developing continually improved models has parallels with the traditional open-source software development approach~\cite{raymond1999cathedral} where the trend is for the community to develop foundational machine learning models like open source software~\cite{raffel2021call,raffel2023building}.

A simple method of merging multiple models is to average the weights of multiple models fine-tuned from the same base initial model. This \textit{model soup} approach~\cite{wortsman2022model} demonstrated significant improvements on relatively large image processing and image classification models. The work pointed to similarity between weighted model averaging leading to a flat local minima. Theoretical~\cite{dziugaite2017computing,kaddour2022flat,petzka2021relative,keskar2017on} and empirical studies dating back from the 1990s~\cite{hochreiter1994simplifying,hochreiter1997flat} demonstrated that flatter local optima generalize better to out-of-distribution shifts. More recent work~\cite{matena2022merging,daheim2024model} examined the theoretical properties and practical implications of weight averaging applied to larger generative models.

\added{
Linear weight averaging is performed as follows. Let $\theta_1, \theta_2 \in \mathbb{R}^d$ represent the weight vectors of two distinct models, where $d$ is the dimension of the weight space. The merged model's weights, denoted as $\theta_\text{new}$, are computed using the following equation: $\theta_\text{new} = \lambda \theta_1 + (1 - \lambda) \theta_2.$
Here, $\lambda \in [0, 1]$ is a weighting parameter that determines the relative contribution of each model to the final merged model. By adjusting $\lambda$, practitioners can control the balance between the two original models' characteristics in the resulting merged model. 
}

Linear weight averaging not only works well for image processing and classification models, but is also effective for image generation models such as latent diffusion models. The release of Stable Diffusion~\cite{rombach2022high} led to a variety of specialist fine-tunes of the model for a variety of different styles, from photorealism, to anime, to realistic painting styles. Popular open-source WebUIs~\cite{automatic1111} enabled \replaced{practitioners}{hobbyists} to experiment with merging different models using linear or spherical linear interpolation (SLERP)~\cite{white2016sampling} of weight parameters of distinct fine-tuned Stable Diffusion models. This has led to the discovery of merged models that combine strengths of specialized models into a single model.
For some time, the most popular Stable Diffusion models were neither the original base models nor the fine-tuned versions, but rather the merged models created by enthusiasts.

\subsubsection{Merging Language Models}
\vskip -0.05in 

Simple weight interpolation works well for merging image generation models that operate in a real valued latent and output space, but to our surprise, they also still work for merging language models under the right conditions, but have performance issues~\cite{ties}. Recently, new methods have been proposed to address merging language models specifically. One method, Task Arithmetic~\cite{ilharco2022editing}, involves building \textit{task vectors} by subtracting pre-trained from fine-tuned model weights, enabling manipulation through arithmetic operations to steer the merged model's behavior. \deleted{The development }

\added{
Task Arithmetic operates as follows.
Let $\theta_\text{base} \in \mathbb{R}^d$ denote the weights of a pre-trained base model.
For each task $k$, let $\theta_k \in \mathbb{R}^d$ represent the weights of a model fine-tuned for that specific task.
The task vector $\tau_k$ for each task $k$ is defined as the difference between the fine-tuned model weights and the base model weights:
$\tau_k = \theta_k - \theta_\text{base}.$
The weights of the new, merged model $\theta_\text{new}$ are then computed by adding a weighted sum of these task vectors to the base model weights: $
\theta_\text{new} = \theta_\text{base} + \sum_{k} \lambda_k \tau_k
$
Here, $\lambda_k$ is a scaling parameter that determines the contribution of each task vector to the final merged model.
}

A key problem of weight interpolation-based methods is that they neglect parameter interference, leading to performance degradation. A recent work~\cite{ties} identifies two key sources of interference: redundant parameter values and conflicting parameter signs across models, and proposes a method that addresses these issues to achieve improved merging performance. The proposed TIES-Merging method addresses information loss in existing merging methods by incorporating three steps: resetting minimal parameter changes, resolving sign conflicts, and merging only aligned parameters. The trimming of redundant parameter changes in TIES-Merging not only resulted in higher performance of merged models, but also achieved better compression for the fine-tuned parameter updates. This compression of fine-tuned weight parameter updates is further investigated in ComPEFT~\cite{yadav2023compeft}.

\added{
The three steps of TIES-Merging are implemented as follows. In the first step, minimal parameter changes are reset by retaining only the most influential components of each task vector $\tau_k$ and setting the rest to zero. This effectively removes redundant or insignificant modifications. The second step resolves sign conflicts across models by determining an aggregate sign for each parameter based on the overall direction of significant changes across all tasks. This ensures that conflicting updates do not cancel each other out. In the final step, only the aligned parameters are merged by selectively averaging the values of parameters that agree with the elected sign from the previous step, while excluding those that disagree. This selective merging process ensures that the final model incorporates only consistent and significant parameter changes from each task-specific model, thereby preserving task-relevant information while minimizing interference.
}

Another recent work~\cite{dare} proposes the DARE method\added{, which} goes further by zeroing out small differences between the fine-tuned model and the original base model, while amplifying the differences. In practice, DARE~\cite{dare} is often used together with Task Arithmetic~\cite{ilharco2022editing} or TIES-Merging~\cite{ties}.
\added{
Specifically, DARE operates as follows. It applies a random mask $m_k \sim \text{Bernoulli}(p)$ to task vector $\tau_k$, where $p$ is the drop rate, and then rescales the remaining parameters:
\begin{equation}
\hat{\tau}_k = \frac1{1 - p} \left( {(1 - m_k) \odot \tau_k} \right).
\end{equation}
The resulting sparsified task vector $\hat{\tau}_k$ is then used to merge models. This task vector can be utilized in the aforementioned linear task arithmetic (referred to as DARE-Linear) or TIES (referred to as DARE-TIES) methods.
}

Model merging is gaining momentum in ML research, and has been applied in areas such as multimodal model development~\cite{aiello2023jointly,sung2023empirical} and merging experts~\cite{aiello2023jointly}. It is also becoming popular in both the broader image generation and LLM communities, where due to the low barrier to entry, many researchers\replaced{ and}{,} engineers\deleted{, and hobbyists alike} are continually merging existing models to create new customized models.

\deleted{Image generation model merging really took off in the user community only after such methods were implemented in open-source toolkits~\cite{automatic1111}.} \added{In the image generation domain, model merging techniques became widely accessible to practitioners especially after their implementation in open-source toolkits~\cite{automatic1111}.} 
Similarly, language model merging really gained popularity after the implementation of \textit{mergekit}~\cite{goddard2024merge,labonne2024merge}. This toolkit provides all of the popular \textit{recipes} for merging language models. In addition to simple linear and spherical interpolation, advanced recipes such as Task Arithmetic, TIES-Merging, DARE are also implemented, offering users the ability to experiment with combining these methods to work with fine-tunes of popular base models such as Mistral~\cite{mistral}. With this toolkit, a large number of capable merged models are developed by the community, and most of the top models on the Open LLM Leaderboard~\cite{openllmleaderboard} are gradually dominated by merged models produced by the community of language model enthusiasts.

Mergekit also introduced an additional method, called \textit{Frankenmerging} which is not based on weight merging, but for the user to experiment with trying to stack different layers from multiple models to sequentially create a new model. This method has the advantage of not tying the user to one particular family of models with a fixed architecture for merging (e.g. Mistral-based models), but potentially creating \textit{new} architectures from merging entirely different models. It is noted that discovering new \textit{Frankenmerging} technique remains a challenge for the community, and requires much more trial and error to discover new recipes for this technique. To date, almost everyone uses a similar \textit{Frankenmerging} merge recipe, and there has been very little trial and error to improve upon it. The field remains highly under-explored, and we believe this is where evolution can help.

\subsubsection{Connection to Evolutionary Neural Architecture Search}
\vskip -0.05in 
Model merging holds significant potential and democratizes the model-building process, making it accessible to a broad range of participants. However, it heavily relies on intuition and domain knowledge. Human intuition, however, has its limits. With the growing diversity of open models and tasks, we need a more systematic approach. We believe evolutionary algorithms, inspired by natural selection, can unlock more effective merging solutions. These algorithms can explore a vast space of possibilities, discovering novel and counter-intuitive combinations that traditional methods and human intuition might miss.

In deep learning, techniques such as Neural Architecture Search (NAS)~\cite{zoph2016neural} employed evolutionary techniques to discover new architectures~\cite{real2019regularized,so2019evolved} that might be non-intuitive for human designers to discover. However, NAS-methods consumed significant computational resources given how each candidate model architecture needed to be trained. Our approach aims to \textit{save} compute resources by discovering structures that take advantage of existing capabilities of existing Transformer blocks that had already consumed considerable resources to train in the first place.

Pre-trained transformer blocks are akin to neural network components one can choose to mix and match using an evolutionary architecture search algorithm. However, unlike NAS, we do not need to train the model, and can evaluate the candidates right away. In a sense, our work resonates with the goals of earlier works of morphology search such as NEAT~\cite{stanley2002evolving} and more recent work such as Weight Agnostic Neural Networks~\cite{gaier2019weight} which evolved neural network structures with certain task-specific inductive biases, without the need to train the weight parameters using gradient descent. Similarly, this was how NAS methods such as SMASH~\cite{stanley2002evolving} avoided costly inner-loop training through the application of a Hypernetwork~\cite{ha2016hypernetworks} to estimate the weights of architectural candidates.

Recent work in language models tried to duplicate the layers of a pre-trained Transformer many times~\cite{sung2023empirical} as a starting point to pre-train a much larger network, to enable efficient training of much larger language models~\cite{young2024yi}. Similarly, in our approach, we are interested in applying automated search methods to discover the effective reuse of existing pre-trained layers.

In this work, we apply evolution to not only automate the optimization of model merging recipes in weight space, confined to a single architecture, but also explore applying evolution to optimize the stacking of layers from different models---a more involved approach that has the potential to create entirely novel neural architectures from existing building blocks. Unlike weight merging optimization which confine space of models to be merged together to fine-tuned descendents of the same parent base model, layer stacking optimization has no such constraints. Our work systematically explores the application of evolution applied in both parameter and layer space.

\section{Methods}
\label{sec:method}

\begin{figure*}[!htb]
\centering
\includegraphics[width=\textwidth]{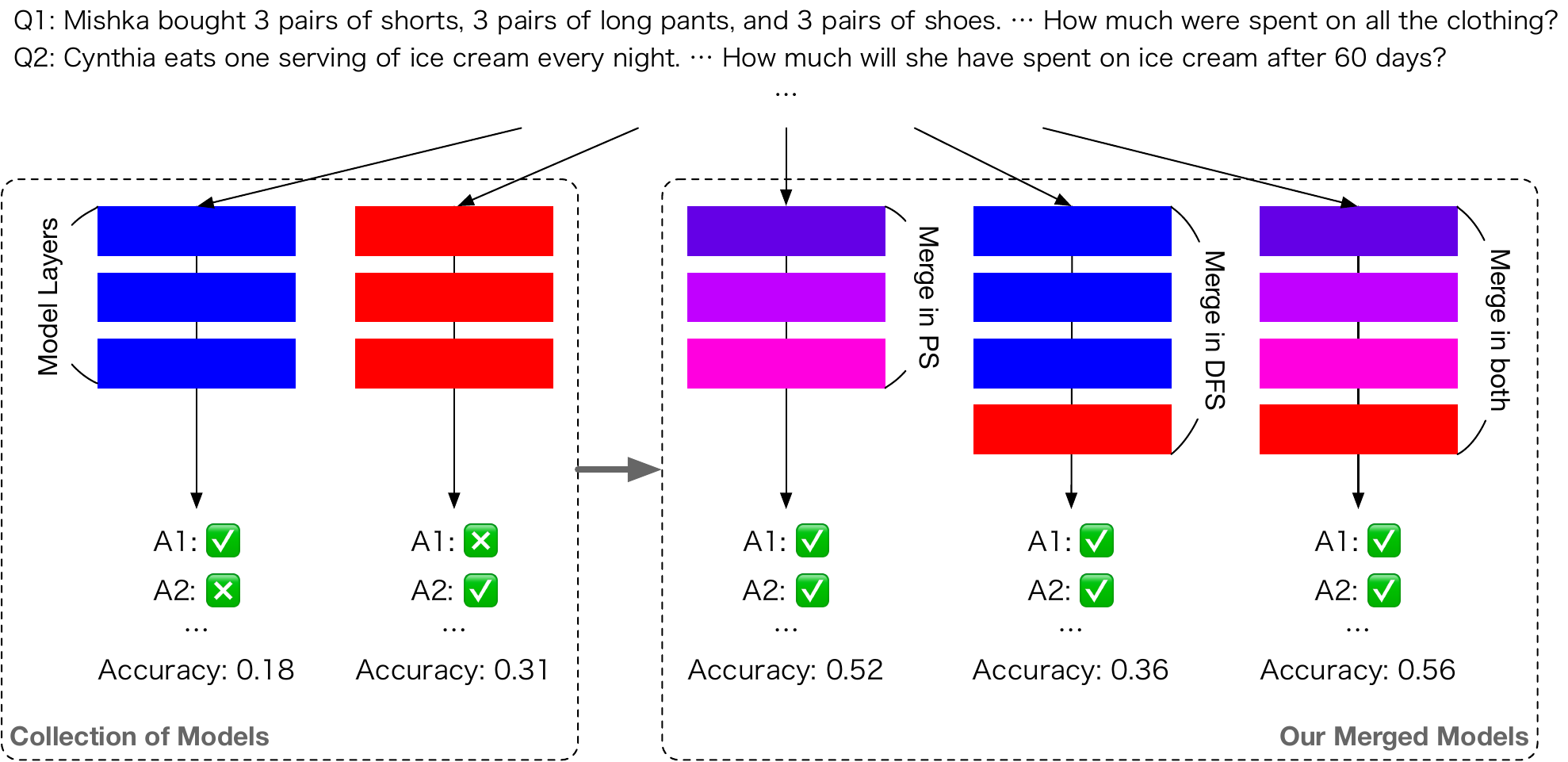}
\vskip -0.05in
\caption{\textbf{Overview of Evolutionary Model Merge}. Our approach encompasses (1) evolving the weights for mixing parameters at each layer in the parameter space (PS); (2) evolving layer permutations in the data flow space (DFS); and (3) an integrated strategy that combines both methods for merging in both PS and DFS. Notice that merging in the PS is not simple copying and stitching of the layers parameters, but also mixes the weights. This merging is akin to blending colors as illustrated here (e.g., red and blue becomes purple). Note that we translated the questions to English for the reader; the models operate on Japanese text.}
\label{fig:overview}
\vskip -0.1in
\end{figure*}

Our goal is to create a unified framework capable of automatically generating a merged model from a selection of foundation models, ensuring that the performance of this merged model surpasses that of any individual in the collection. Central to our approach is the application of evolutionary algorithms, which we employ to refine the intricacies involved in model merging. To systematically address this challenge, we first dissect the merging process into two distinct, orthogonal configuration spaces, analyzing their individual impacts. Building on this analysis, we then introduce a cohesive framework that seamlessly integrates these spaces. Figure~\ref{fig:overview} provides a schematic representation of our approach.

\subsection{Merging in the Parameter Space}

Model merging in the parameter space (PS) aims to integrate the weights of multiple foundational models into a unified entity with the same neural network architecture, yet outperforming the individual models. While various strategies for combining model parameters exist~\cite{matena2022merging,wortsman2022model}, our approach leverages task vectors analysis to understand each model's strengths, based on the specific tasks they are optimized for or excel in~\cite{ilharco2022editing}. Specifically, we enhance TIES-Merging with DARE~\cite{ties,dare}, allowing for more granular, layer-wise merging (In this paper, by ``layer'' we mean the input/output embedding layers or a transformer block). We establish merging configuration parameters for sparsification and weight mixing at each layer, including input and output embeddings. These configurations are then optimized using an evolutionary algorithm, such as CMA-ES~\cite{hansen2006cma}, for selected tasks, guided by critical task-specific metrics (e.g., accuracy for MGSM, ROUGE score for VQA).

\subsection{Merging in the Data Flow Space}
\label{sec:dfs_merging}

Recent analysis and discoveries imply that knowledge is stored distributedly in language models~\cite{geva2022transformer,lesswrong2021interpreting,meng2022locating}, suggesting simple yet novel model merging possibilities in the data flow space (DFS). Unlike merging in PS, model merging in DFS preserves the original weights of each layer intact. Instead, it optimizes the inference path that tokens follow as they traverse through the neural network. For example, after the $i$-th layer in model $A$, a token may be directed to the $j$-th layer in model $B$.

In our initial effort in this domain, we limit ourselves to serial connections and non-adaptive configurations, deferring the investigation of more flexible model merging to future work. Concretely, with a collection of $N$ models and a budget $T$, our method searches for a sequence of layer indices $L_{i,j}^{(t)}$ that delineates the path all the tokens should follow for a specific task. Here $L_{i,j}$ denotes the $j$-th layer in the $i$-th model, with $t \in [1, T]$ marking the step in the inference path.

One can quickly imagine how large the search space is. Assuming the total number of layers across all models is $M$, the size of the search space is $(M+1)^T$, here the extra one indicates the inclusion of a pass-through layer. Even with a modest setting of $M=64$ (e.g., 2 models of 32 layers each) and $T=60$, this setting translates to an astronomically large search space, a challenge even for a capable evolutionary search algorithm. Luckily, our preliminary studies indicated that certain layer arrangements, particularly repetitive or permuted sequences from earlier in the model, can adversely affect performance. Based on this, we modify our settings to include an indicator array $\mathcal{I}$ of size $T=M \times r$ in the evolutionary search space, here $r$ is number of repetitions.

Conceptually, we layout all the layers in sequential order (i.e., all layers in the $i$-th model followed by those in the $i+1$-th model) and repeat them $r$ times, the indicator array then manages the inclusion/exclusion of layers. If $\mathcal{I}_i > 0$ we include the layer corresponding to index $i$ in the slots in the merged model, otherwise we exclude it. Consequently, our search space is reduced to $2^T$, which is still large, but tractable for evolutionary search.

In our search, we only optimize the data inference path inside the merged model and keep parameters in the models intact. In this setting, a layer may face an input whose distribution is different from what it is used to (from its original model), leading to unexpected outputs. For example, \added{our preliminary studies~\cite{sun2024transformer}} show that swapping a pair of neighboring layers in a language model makes its performance drop. Although more theoretical studies are needed to model the distribution shift, empirically we find that appropriately scaling an input that wishes to go from layer $i$ to $j$ by $W_{ij}$ help alleviate the problem. Here, $W \in \mathcal{R}^{M \times M}$ is a matrix that is also optimized by the evolutionary search together with the indicator array $\mathcal{I}$.

The size of $W$ grows quadratically with $M$, for scenarios involving a large number of layers. An alternative approach to contain the search space size involves parameterizing $W$ with a neural network~\cite{schmidhuber1992learning,ha2016hypernetworks}. We can instead evolve a feed-forward network to output the scaling weights conditioned on the layer and step indices: $W_{ij}=\pi_{\theta}(i, j, t)$ where $\theta$'s are the parameters to be evolved, whose size does not change when $M$ grows.

\subsection{Merging in Both Spaces}

Model merging in the PS and in the DFS are orthogonal approaches, however, it is straightforward to combine these disentangled methods and further boost the performance of a merged model. As we show in the rightmost illustration in Figure~\ref{fig:overview} and in Section~\ref{sec:japanese_math}, it is possible to first apply PS merging to a collection of models, and then put back this merged model in the collection and apply DFS merging from this enlarged collection.

This can be extremely helpful when one considers model merging with multiple objectives, wherein PS merging can be applied first to produce several merged models each of which targets one of the muliple objectives of interest, and then DFS merging is applied with multi-objective genetic algorithms such as NSGA-II~\cite{deb2002fast} to further expand the final model's performance in relevant metrics.

\section{Results}

Most merged models in the community optimize for a narrow set of tasks defined in The Open LLM Leaderboard~\cite{openllmleaderboard}. Our motivation is for evolutionary search to discover novel ways to merge different models from vastly different domains (e.g., non-English language and Math, or non-English language and Vision) which might be difficult for human experts to discover effective merging solutions themselves. Furthermore, effectively merging models from very different domains can lead to models of wider real-world applicability and enable us to develop models beyond the large population of models that are optimized for the narrow range of tasks defined by a leaderboard.

We demonstrate our Evolutionary Model Merge approach described in Section~\ref{sec:method} by evolving a Japanese LLM capable of Math reasoning, and a Japanese VLM proficient in handling culturally-specific content.
Specifically, in Section~\ref{sec:japanese_math}, we apply evolution to merge a Japanese LLM with an English Math LLM to build a Japanese Math LLM, and in Section~\ref{sec:experiments/vlm}, we apply evolution to merge a Japanese LLM with an English VLM to create a Japanese VLM.

\subsection{Evolving Japanese Math LLM}
\label{sec:japanese_math}

\subsubsection{Experimental Setup}
\paragraph{Source Models}
To develop a model capable of solving math problems in Japanese, we apply evolutionary model merge on a set of source models containing a Japanese LLM and Math LLMs: \texttt{shisa-gamma-7b-v1}~\cite{shisagamma7b} (Japanese LLM), \texttt{WizardMath-7B-V1.1}~\cite{wizardmath} and \texttt{Abel-7B-002}~\cite{abel}. All these models are fine-tuned from \texttt{Mistral-7B-v0.1}~\cite{mistral}.

\paragraph{Dataset}

For testing, we used the MGSM dataset~\cite{mgsm}, a multilingual translation of a subset of the GSM8k dataset~\cite{gsm8k}. The Japanese test set of MGSM, consisting of 250 samples, was used for the final evaluation. \added{Specifically, MGSM contains translations of the first 250 samples (ID 0-249) from the GSM8k test set.} We used a different dataset for evolutionary search to avoid overfitting the test set. Specifically, we translated the remaining 1069 samples (out of 1319 examples) of the GSM8k test set that were not included in the MGSM test set into Japanese. \added{These correspond to samples with IDs 250-1318 in the original GSM8k test set, ensuring no overlap with the MGSM Japanese test set.} One may think that it is more natural to translate the GSM8k training set. However, in our preliminary efforts, this approach did not work well. Since open-source math models were trained on the GSM8k training set, we were unable to perform accurate evaluations.

\paragraph{Evaluation} 
We evaluated the ability to generate Japanese answers to Japanese math problems.
Therefore, we considered an answer correct if it met the following criteria: (1) the concluding numerical value must be correct, and (2) the reasoning text should be written in Japanese.

We treated the last numerical value appearing in the output as the answer. We needed to adopt this heuristic because we are merging multiple models that were trained in different formats, which made it difficult to correct the output format. This method appeared to extract the answers correctly in almost all cases. Additionally, to determine the language of the output, we utilized fasttext~\cite{joulin2016bag, joulin2016fasttext}. We used greedy sampling for generation and calculated the zero-shot pass@1 accuracy.

\paragraph{Optimization}
For optimization in PS, we used the CMA-ES\cite{hansen2006cma} algorithm implemented in Optuna~\cite{optuna}  with default hyper-parameters. Specifically, we set all initial parameter values to 0.5, sigma to 1/6, and the population size to $4 + \lfloor 3  \ln(n_{\text{params}}) \rfloor$, where $n_{\text{params}}$ is the number of parameters to optimize.
The fitness value is defined as the accuracy for all 1069 training samples. Please note that this set is disjoint from MGSM's test set.
The optimization was conducted for 1000 trials, and the best trial with respect to the training accuracy was chosen as the final model.
We decided to employ TIES-Merging~\cite{ties} with DARE~\cite{dare} through preliminary experiments, and optimized its parameters.

In our DFS merging experiments, $M=64, r=3$, and consequently, $T=M \times r = 192$.
We kept the last 200 examples in the training data as our validation set and optimize on the rest of the data with a batch size of 200. We report the performance of the snapshot that achieved the highest accuracy in the validation set, and the test set is strictly isolated from the optimization process.
We adopted CMA-ES in EvoJAX~\cite{evojax2022}, it optimized $\mathcal{I}$ and $W$ for a total of 100 generations with a popluation size of 128, and we used the default hyper-parameters.
We limited our DFS merging to two models $A$ and $B$ to ensure that the final model remains modest in size and can be run on a single GPU, but in principle, the methodology can scale to merging multiple models.
During the merging, model $A$'s tokenizer, and input/output embeddings are utilized. Furthermore, to maintain compatibility with the embedding layers, we mandate that the initial and final transformer layers of model $A$ define the start and the end of the inference path. We initialized the indicator array $\mathcal{I}$ so that all layers in model $A$ are more likely to be included as initial hops in the inference path to shorten the search time.

\subsubsection{Experimental Results}

Table~\ref{table:math} summarizes the performance of the LLMs on Japanese Math and overall Japanese benchmark tasks. The MGSM-JA column reports the results from the MGSM test set, employing the previously described metrics. The Japanese language model (Model 1) demonstrates limited mathematical proficiency, while the Math models (Models 2 and 3), though mathematically adept, show insufficient command of the Japanese language. Consequently, all three models score low scores on the MGSM-JA, with accuracy at or below 30.0.

\begin{table}[!htb]
\small
\centering
\caption{\textbf{Performance Comparison of the LLMs.} Models 1--3 are source models, Models 4--6 are our optimized merge models, and Models 7--11 are provided for reference. MGSM-JA measures math ability in Japanese, and JP-LMEH evaluates general Japanese language capabilities, with a detailed breakdown provided in Table~\ref{table:lm-eval-harness}.}
\vskip 0.05in 
\label{table:math}
\begin{tabular}{rllrrr}
\toprule
\textbf{\#} & \textbf{Model} & \textbf{Type} & \textbf{Size} & 
\textbf{MGSM-JA} {\footnotesize (acc $\uparrow$) } &
\textbf{JP-LMEH} {\footnotesize (avg $\uparrow$) } \\
 \midrule
 0 & Mistral-7B-v0.1 & EN general & 7B & 2.0 & 55.7 \\
 1 & Shisa Gamma 7B v1 & JA general & 7B & 9.6 & 66.1 \\
2 & WizardMath 7B v1.1 & EN math & 7B & 18.4 & 60.1 \\
3 & Abel 7B 002 & EN math & 7B & 30.0 & 56.5 \\
\midrule
4 & Ours (PS) & 1 + 2 + 3 & 7B & 52.0 & 70.5 \\
5 & Ours (DFS) & 3 + 1 & 10B & 36.4 & 53.2 \\
6 & Ours (PS+DFS) & 4 + 1 & 10B & 55.2 & 66.2 \\
\midrule
7 & Llama 2 70B & EN general & 70B & 18.0 & 64.5 \\
8 & Japanese StableLM 70B & JA general & 70B & 17.2 & 68.3 \\
9 & Swallow 70B & JA general & 70B & 13.6 & \textbf{71.5 }\\
10 & GPT-3.5 & commercial & - & 50.4 & - \\
11 & GPT-4 & commercial & - & \textbf{78.8} & - \\
\bottomrule
\end{tabular}
\end{table}

In contrast, our merged models (Models 4--6) manifest a substantial elevation in performance. Notably, the model merged in PS (Model 4) achieves an impressive score of 52.0, highlighting the remarkable potential in combining models with distinct expertises. The DFS-merged model (Model 5) also shows a performance enhancement, with an over 6 percent increase in accuracy compared to the source models. While the leap in performance is not as pronounced as with PS merging, it still proves to be a valid and orthogonal approach. Finally, our hybrid model (Model 6), which integrates both merging strategies, shows further enhancements on the task. \added{The order of the source models in the indicator array $\mathcal{I}$ affects the performance of the DFS merging method. We conduct experiments with all possible combinations and report the best scores in Table~\ref{table:math}, we defer detailed analysis to Sec~\ref{sec:dfs_analysis}.}

\begin{figure}[!ht]
\centering
\includegraphics[width=1.0\textwidth]{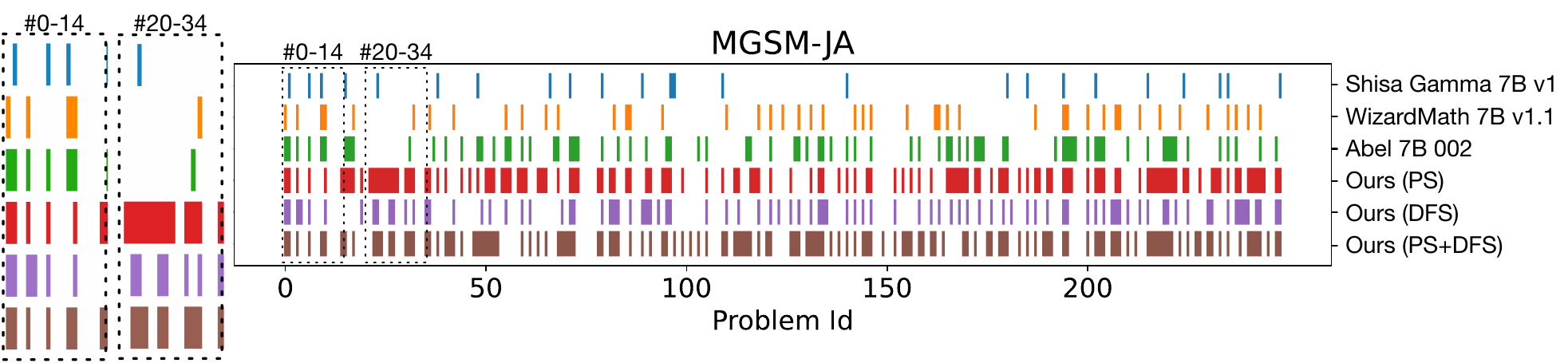}
\vskip -0.1in
\caption{\textbf{Performance Overview.} The figure depicts the success of various models on the MGSM-JA task, with each of the 250 test problems represented along the x-axis by problem ID. Correct answers are indicated by colored markers at the corresponding positions.}
\label{fig:macro_analysis}
\end{figure}

Figure~\ref{fig:macro_analysis} gives an overview of the five models' ``answer sheet'' on the math problems. Our merged models retain the foundational knowledge in the source models, as evidenced by the similar score patterns on problems 1--15. Moreover, they exhibit emergent capabilities, successfully tackling problems that stumped the source models (e.g., problems 20--30). Evidently, by effectively integrating a Japanese LLM and mathematical models, we have succeeded in producing models that are proficient in both Japanese language understanding and mathematical problem-solving.

Furthermore, Table~\ref{table:lm-eval-harness} presents the results of evaluating the general Japanese language ability using the Japanese Language Model Evaluation Harness (JP-LMEH) benchmark suite. This benchmark suite consists of nine tasks, and the average score across these tasks is widely used as an indicator of overall Japanese language proficiency. Our models achieve remarkably high scores of 70.5 and 66.2, surpassing the source models and even the previous state-of-the-art 70B parameter Japanese LLM (specifically, Japanese StableLM 70B), despite having only 7B to 10B parameters.

Compared to the source Japanese model (\texttt{shisa-gamma-7b-v1}), it is interesting to note that there is not only a significant improvement in the MGSM scores but also improvements in other tasks such as JSQuAD and JAQKET. \added{However, we also acknowledge that there are certain tasks, such as JNLI and MARC (for DFS), where performance has dropped.} It should be noted that the MGSM scores here do not match those in Table~\ref{table:math}, due to the differences in evaluation protocols (few-shot, prompting, etc.). Additional details are provided in Appendix~\ref{sec:appendix-eval}, including extensive comparisons with other models in Table~\ref{table:lm-eval-harness-full}.

Moreover, Appendix~\ref{sec:case-study} showcases intriguing examples that demonstrate the utility of our models merged using evolution. The merged models correctly answered questions that require both knowledge about Japanese-culture \textit{and} Math ability. In contrast, even if such Japanese questions were translated into English and answered in English, English math models would likely fail to provide the correct answers as they may not be aware of Japanese culture-specific context in the questions.

\begin{table}[!htb]
\centering
{\fontsize{8pt}{8pt}\selectfont
\setlength{\tabcolsep}{2.5pt}
\caption{\textbf{Breakdown of JP-LMEH Scores for Japanese Language Proficiency}. JP-LMEH (Japanese Language Model Evaluation Harness) is a benchmark suite consisting of 9 tasks, and the average score (Avg column) is used as an indicator of overall Japanese language proficiency. Full results are in Table~\ref{table:lm-eval-harness-full}}
\vskip 0.05in 
\label{table:lm-eval-harness}
\begin{tabular}{lrrrrrrrrrrr}
\toprule
  &  & \multicolumn{10}{c}{\textbf{JP Language Model Evaluation Harness}} \\
\textbf{Model} &\textbf{Size} & JComQA & JNLI & MARC & JSQuAD & JAQKET & XLSum & XWino & MGSM & JCoLA & \textbf{Avg} \\
\midrule
Shisa Gamma 7b v1 & 7B & 91.2 & \textbf{72.1} & 94.6 & 73.9 & 68.0 & \textbf{25.9} & 80.5 & 29.6 & 58.7 & 66.1 \\
WizardMath 7B V1.1 & 7B & 74.7 & 42.7 & 90.4 & 84.6 & 68.5 & 22.3 & 69.8 & 38.8 & 48.9 & 60.1 \\
Abel 7B 002 & 7B & 70.3 & 51.8 & 62.3 & 83.8 & 69.0 & 22.5 & 68.2 & 28.0 & 52.7 & 56.5 \\
\midrule
Ours (PS) & 7B & 89.1 & 65.7 & \textbf{95.4} & 89.5 & 77.7 & 25.5 & 81.2 & \textbf{50.0} & 60.5 & 70.5 \\
Ours (DFS) & 10B &  67.7 & 58.2 & 53.5 & 66.8 & 54.3 & 17.3 & 65.6 & 30.0 & 65.6 & 53.2 \\
Ours (PS+DFS) & 10B &  88.2 & 50.3 & 91.5 & 78.6 & 77.8 & 23.2 & 73.0 & 40.0 & \textbf{73.0} & 66.2 \\
\midrule
Llama 2 70B & 70B & 80.2 & 53.4 & 94.4 & 91.6 & 80.1 & 21.8 & 73.6 & 30.4 & 54.6 & 64.5 \\
Japanese Stable LM 70B & 70B & 91.2 & 50.4 & 92.9 & 87.1 & 88.4 & 24.3 & 82.0 & 37.2 & 61.7 & 68.3 \\
Swallow 70B & 70B & \textbf{95.3} & 57.2 & 91.7 & \textbf{94.1} & \textbf{93.9} & 23.1 & \textbf{83.3} & 45.2 & 59.5 & \textbf{71.5} \\
\bottomrule
\end{tabular}
}
\vskip -0.1in
\end{table}

\subsubsection{Analysis}
\label{sec:llm_analysis}

\begin{figure}[!ht]
  \centering
    \vskip -0.15in
    \includegraphics[width=0.4\linewidth]{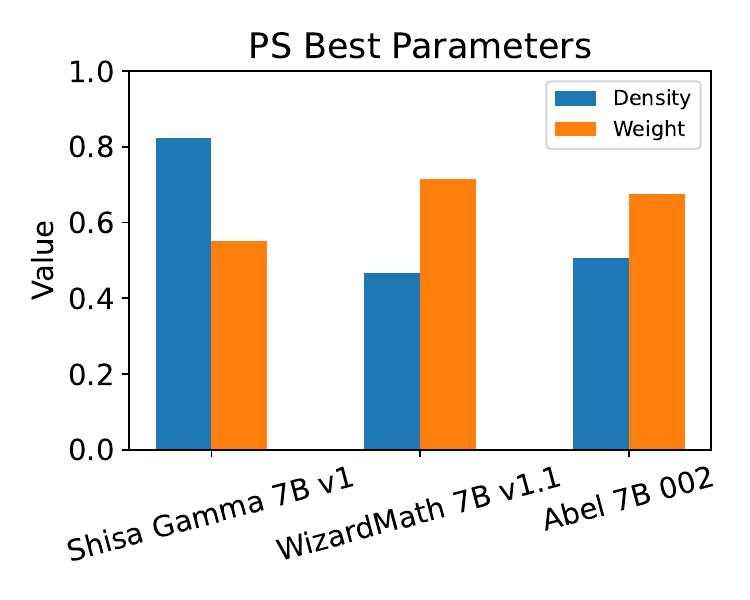}
    \vskip -0.15in
    \caption{\textbf{Evolved Configurations for PS merging.} Although the weights are similar across the 3 source models, the pronounced density from the Japanese LLM underscores its pivotal role.}
    \label{fig:ps_analysis}
\end{figure}

In our exploration of model merging in the parameter space, we experimented with diverse configurations, such as varying the assignment of merging parameters across different layer groups. However, due to a constrained dataset, we did not witness notable improvements in performance correlating with increase in configuration complexity. Consequently, we focused our reporting on a PS merged model (Model 4 in Table~\ref{table:math}) that adopts the simplest setting: considering each source model as a singular layer and allocating two DARE-TIES associated parameters to each for evolutionary merging. Figure~\ref{fig:ps_analysis} illustrates the evolved parameter configuration post PS merging.

The CMA-ES optimization results reveals that all three models are important, as suggested by the uniformity of the optimized weighting values. 
\added{The fact that the sum of the weights exceeds 1 and is approaching 2, is noteworthy. This suggests that a combination method which amplifies the contributions of the models, rather than a simple interpolation, proved to be more effective.}

The dominant density from the Japanese LLM suggests its critical contribution to solving the task. We conjecture that this may also be partially attributed to the Japanese LM's larger amount of fine-tunings from the Mistral base model.
\added{Japanese LMs based on English models, such as \texttt{Shisa-Gamma-7B-v1}, are typically created through a two-step process: continued pretraining and instruction fine-tuning. The continued pretraining phase involves learning from a significantly larger dataset compared to standard fine-tuning. For instance, \texttt{Shisa-Gamma-7B-v1} is based on a Japanese base model that underwent continued pretraining on 100B tokens of Japanese text. The resulting differences in weights between \texttt{Shisa-Gamma-7B-v1} and the original \texttt{Mistral-7B-v1} likely encapsulate more information than standard fine-tuning, making them more challenging to sparsify.}
In line with the discussion in~\cite[Section~4.6]{dare}, the sparsification of DARE tends to degrade performance when applied to such extensively fine-tuned models. Our evolutionary search has seemingly managed to address this issue by increasing the density for the Japanese LM.

The parameter settings derived from our PS merging experiments align well with outcomes from our DFS merging efforts. By incorporating the PS-merged model into our pool of source models and applying DFS merging across all potential pairings, we observed optimal performance with the combination of the PS-merged model and the Japanese language mode (Model 6 in Table~\ref{table:math}). This finding echoes the significant influence of the Japanese language model, as indicated by its notable presence in Figure~\ref{fig:ps_analysis}, and reaffirms the substantial promise of evolutionary model merging.

\begin{figure}[!ht]
  \centering
    \vskip -0.15in
    \includegraphics[width=\linewidth]{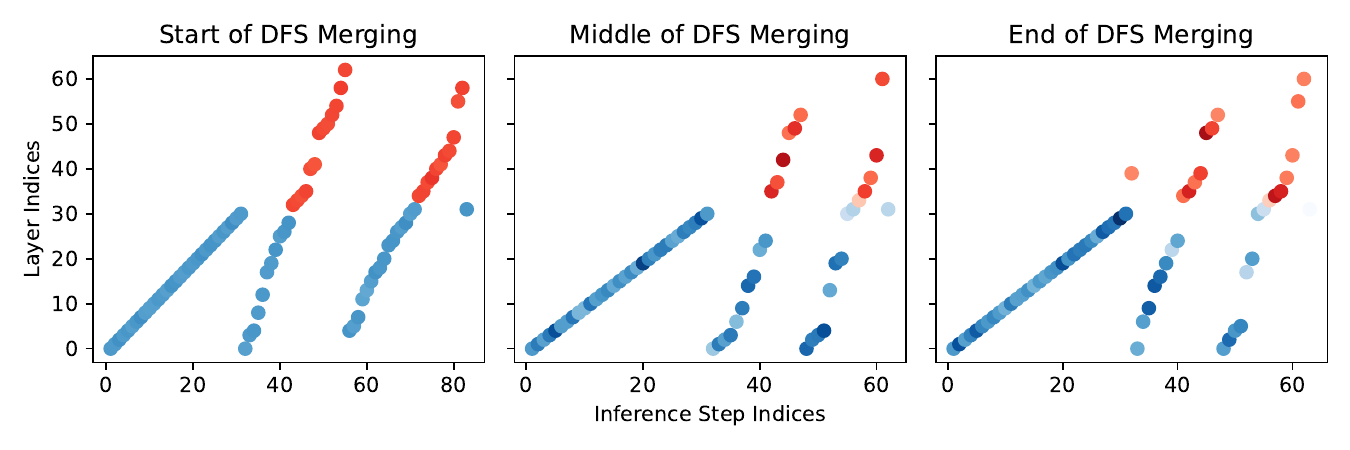}
    \vskip -0.15in
    \caption{\textbf{Evolved Configurations for DFS Merging of models $A$ and $B$.} The 3 figures depict the evolution of the inference path on the MGSM-JA task. The y-axis represents the layer index $l \in [1, M]$, and the x-axis corresponds to the path index $t \in [1, T]$. Blue markers indicate path steps utilizing layers from model $A$, red markers denotes those from $B$. \deleted{Marker size reflects the magnitude of the scaling factor $W_{ij}$.} \added{Marker color intensity reflects the magnitude of the scaling factor $W_{ij}$.} The evolutionary search result includes most layers in $A$ at an early stage and then alternates between layers from both models. Result from our 10B model (PS+DFS).} 
    \label{fig:dfs_analysis}

\end{figure}

Figure~\ref{fig:dfs_analysis} displays the evolution of the inference path, where our approach consistently recognized the value of initial steps, incorporating every layer from the first model (our PS-merged model), except for the last decoding layer and the embedding layer. As the process advanced, the method refined the selection to a smaller, more effective set of layers and strategically alternated between layers from both contributing models. Significantly, the scaling parameters $W_{ij}$ emerged as crucial elements, our ablation studies revealed that eliminating them in the evolved model (e.g., by setting $W_{ij}=1$) led to a performance decline exceeding 20 percent, highlighting their importance in the model's efficacy.

\subsection{Method Behavior Analysis}
\label{sec:method_behavior_analysis}

\added{
In this section, we conduct a more in-depth analysis of our method's performance. We compare it with multiple types of baselines, experimentally analyze the effect of manual model selection, verify the scalability of our proposed approach to larger models, and show results from ablation studies.
}

\subsubsection{Comparison with Unoptimized Model Merging}

\added{
Table~\ref{table:math-baselines} presents the performance of models created through unoptimized model merging. We utilized three methods: TIES-Merge~\cite{ties}, DARE-TIES~\cite{dare}, and Frankenmerging~\cite{goddard2024merge}.
We used mergekit to establish the baselines.
Since merkit does not include default hyperparamters, we adopted those from a widely recognized resource for learning about model merging and mergekit~\cite{labonne2024merge}.
}

\begin{table}[!htb]
\small
\centering
\caption{\textbf{Performance Comparison of Proposed Methods and Baselines.} Results show MGSM-JA accuracy and JP-LMEH average scores for our proposed methods (PS, DFS, PS+DFS), unoptimized model merging techniques, fine-tuning approaches, and experiments with irrelevant model distraction.}
\vskip 0.05in 
\label{table:math-baselines}
\begin{tabular}{lrrr}
\toprule
\textbf{Method} & \textbf{Size} & 
\textbf{MGSM-JA} {\footnotesize (acc $\uparrow$) } &
\textbf{JP-LMEH} {\footnotesize (avg $\uparrow$) } \\
\midrule
Ours (PS) & 7B & 52.0 & \textbf{70.5} \\
Ours (DFS)  & 10B & 36.4 & 53.2 \\
Ours (PS+DFS)  & 10B & \textbf{55.2} & 66.2 \\
\midrule
\multicolumn{4}{l}{\textbf{\textit{Unoptimized model merging:}}} \\
TIES-Merge~\cite{ties}  & 7B & 4.4 & 63.7 \\
DARE-TIES~\cite{dare} & 7B & 35.2 & 66.3 \\
Frankenmerging~\cite{goddard2024merge} & 9B & 0.0 & 16.1 \\
\midrule
\multicolumn{4}{l}{\textbf{\textit{Fine-tuning:}}} \\
LoRA, Shisa Gamma 7B v1 & 7B & 20.8 & 66.4 \\
LoRA, WizardMath 7B v1.1  & 7B & 43.2 & 55.9 \\
LoRA, Abel 7B 002 & 7B & 40.0 & 52.8 \\
Full, Shisa Gamma 7B v1  & 7B & 19.2 & 54.5 \\
Full, WizardMath 7B v1.1  & 7B & 33.6 & 43.4 \\
Full, Abel 7B 002  & 7B & 36.4 & 37.5 \\
\midrule
\textbf{\textit{Distraction:}} \\
$+0$ irrelevant model & 7B & 50.0 & 65.9 \\
$+1$ irrelevant model & 7B & 46.8 & 64.2 \\
$+2$ irrelevant model & 7B & 46.8 & 64.1 \\
$+4$ irrelevant model & 7B & 48.4 & 64.0 \\
$+8$ irrelevant model & 7B & 40.8 & 65.8 \\
\bottomrule
\end{tabular}
\end{table}

\added{
For the PS baseline equivalents (TIES-Merge and DARE-TIES), we used \texttt{shisa-gamma-7b-v1}, \texttt{WizardMath-7B-V1.1}, and \texttt{Abel-7B-002} as the source models, allowing direct comparison with our PS experiments. Similarly, for the DFS baseline equivalent (Frankenmerging), we used the same combination of source models as in our DFS experiments, enabling direct comparisons.
}

\added{
The results indicate that TIES-Merge and Frankenmerging led to decreased performance in both MGSM-JA and JP-LMEH, suggesting unsuccessful merging. Notably, Frankenmerging resulted in a MGSM-JA score of 0, which is consistent with its general tendency to often produce models that fail to generate meaningful outputs. DARE-TIES performed relatively better among these baselines, with a slight improvement in MGSM-JA compared to the source models. However, our PS model significantly outperformed all these baselines, and the hybrid merging method (PS+DFS) gives the best test performance on the target task.
}

\subsubsection{Comparison with Fine-tuning}

\added{
Table~\ref{table:math-baselines} also compares our evolutionary model merging approach with the fine-tuning method. To ensure a fair comparison, we used the identical dataset. Specifically, we utilized the 1069 samples of Japanese-translated GSM8k that were employed in our PS and DFS optimization for fine-tuning the models. We tested learning rates of 1e-5, 5e-5, and 1e-4, choosing the optimal one. We set the number of epochs to 3 and tried both LoRA and full parameter fine-tuning. We experimented with all three source models: \texttt{shisa-gamma-7b-v1}, \texttt{WizardMath-7B-V1.1}, and \texttt{Abel-7B-002}.
}

\added{
The MGSM-JA scores show that while fine-tuning generally improved performance over the original models, the highest score achieved was 43.2. This is notably lower than our proposed method, which exceeded a score of 50. Additionally, unlike our approach, fine-tuning often led to significant decreases in JP-LMEH scores. This suggests that fine-tuning may be more data-hungry and prone to catastrophic forgetting issues.
}

\added{
\textbf{Other potential fine-tuning baselines.}
Another potential fine-tuning baseline would be combining all data used by all source models to fine-tune a base model. However, this approach is often impractical in general settings, and in our specific case, it is indeed impossible to implement for comparison. We elaborate on these points below. Our proposed approach's strength lies in its ability to freely combine openly published models from platforms like HuggingFace. However, model publishers often only release the models themselves, not the training datasets or detailed training methods, as these are considered proprietary knowledge. For instance, the training data for \texttt{WizardMath-7B-V1.1} and \texttt{Abel-7B-002} is not publicly available. \texttt{shisa-gamma-7b-v1} is a further fine-tuned version of \texttt{japanese-stablelm-base-gamma-7b} (which is \texttt{Mistral-7B-v0.1} fine-tuned for Japanese), and the data used for training japanese-stablelm-base-gamma-7b is also not public. Moreover, the Japanese fine-tuning for \texttt{japanese-stablelm-base-gamma-7b} involved 100B tokens, indicating a substantial computational cost (such as more than 10,000 GPU hours). Therefore, a comparison with this approach is not feasible. It is also worth noting that such fine-tuning would require careful tuning of dataset mixing ratios, which is non-trivial and costly.
}

\subsubsection{Impact of Manual Model Selection}

\added{
In our main experiments, we manually selected three source models. To assess the broader applicability of our approach, it is useful to investigate how sensitive the method is to the selection of source models. To this end, we conducted experiments to measure the distraction effect of including irrelevant models, which simulate the scenario where suboptimal models are inadvertently selected.
}

\added{
To investigate the sensitivity of our method to source model selection, we conducted experiments to measure the distraction effect of including irrelevant models. In addition to our original seed models (\texttt{shisa-gamma-7b-v1}, \texttt{WizardMath-7B-V1.1}, \texttt{Abel-7B-002}), we experimented with one, two, four, and eight irrelevant models, and performed PS experiments to examine the impact on performance.
The irrelevant models, chosen for their lack of relation to Japanese or mathematics, were: \texttt{leo-mistral-hessianai-7b} (German), \texttt{lince-mistral-7b-it-es} (Spanish), \texttt{Mistral-7B-v0.3-Chinese-Chat} (Chinese), \texttt{BioMistral-7B} (biomedical domain), \texttt{Claire-Mistral-7B-0.1} (French), \texttt{komt-mistral-7b-v1} (Korean), \texttt{Mistral-7B-v0.1-arabic} (Arabic), and \texttt{Loquace-7B-Mistral} (Italy). All these models are openly available on HuggingFace.
}

\added{
We slightly modified the PS optimization by adding weight normalization, which became necessary due to the increased number of models (up to 11 in total). This modification explains why the "+0 irrelevant model" result does not exactly match our original PS result.
}

\added{
The results, shown in the "Distraction" part of Table~\ref{table:math-baselines}, are generally favorable. As expected, using only the initial three models yielded the best results. However, the performance degradation when adding irrelevant models was limited, especially up to four additional models, and even with 8 additional models, the performance did not collapse. This demonstrates that our optimization method is relatively robust to manual source model selection. Due to the huge GPU memory requirement at inference time, the distraction experiments were not conducted for the DFS merging method. We instead present behavior analysis and ablation studies on the topic in the following sections. 
}

\subsubsection{Scaling to Larger Models}

\added{
Finally, we demonstrate our proposed methods also work with larger models. Concretely, we used fine-tuned versions of \texttt{Llama-2-13b}: \texttt{ELYZA-japanese-Llama-2-13b-instruct} as the Japanese general model and \texttt{MetaMath-13B-V1.0} as the English math model. All other settings remained identical to the 7B experiments. Due to the limited availability of suitable 13B models, we could only select two source models.
}

\added{
The results are shown in Table~\ref{table:13b}. Similar to the 7B case, the MGSM-JA score significantly improved over both source models. This consistent behavior at the 13B size demonstrates the scalability of our evolutionary model merging approach.
}

\begin{table}[!htb]
\small
\centering
\caption{\textbf{Scaling to 13B Models with ablation studies on DFS merging.} Performance comparison and ablation study results on MGSM-JA and JP-LMEH tasks. Rows in gray are ablation studies.}
\vskip 0.05in 
\label{table:13b}
\begin{tabular}{rlllrr}
\toprule
\textbf{\#} & \textbf{Model} & \textbf{Type} &
\textbf{Size} & 
\textbf{MGSM-JA} {\footnotesize (acc)} &
\textbf{JP-LMEH} {\footnotesize (acc)} \\
\midrule
0 & Llama-2-13b & EN general & 13B & 2.8 & 54.1 \\
1 & ELYZA-japanese-Llama-2-13b-instruct & JA general & 13B & 13.2 & 60.2 \\
2 & MetaMath-13B-V1.0 & EN math & 13B & 8.0 &  48.7 \\
\midrule
3 & Ours (PS) & 1 + 2 & 13B & 31.2 & 59.7 \\
\rowcolor{lightgray}
4 & Ours (DFS) & 1 + 2 & 19B & 12.8 & \textbf{61.1} \\
5 & Ours (DFS) & 2 + 1 & 21B & 23.2 & 46.6 \\
\rowcolor{lightgray}
6 & Ours (PS+DFS w/o $W$) & 3 + 1 & 19B & 26.4 & 58.1 \\
7 & Ours (PS+DFS) & 3 + 1 & 21B & \textbf{34.0} & 60.4 \\
\bottomrule
\end{tabular}
\end{table}

\added{
The 7B models achieved higher MGSM-JA scores (52.0, 36.4, 55.2) than the 13B models (31.2, 23.2, 34.0) for (PS, DFS, PS+DFS). This difference is attributed to the source models rather than our proposed method. \texttt{Mistral-7B-v0.1} is known to outperform \texttt{Llama-2-13b} in performance, especially in basic mathematical abilities, despite its smaller size. This is reflected in the performance of math-specialized models: \texttt{WizardMath-7B-V1.1} scores 83.2 on GSM8k (an English math benchmark), while \texttt{MetaMath-13B-V1.0} scores 72.3. This difference explains the performance gap in Japanese math abilities between our merged 7B and 13B models.
}

\added{
On the other hand, our DFS and PS+DFS models (models \#4 and \#7) achieved higher JP-LMEH average scores than \texttt{ELYZA-japanese-Llama-2-13b-instruct}, the Japanese general model.
In general, the qualitative behavior on JP-LMEH is consistent between 7B and 13B models. Both sets of results show improvements in JSQuAD, JAQKET, and MGSM scores, with slight decreases in JNLI and MARC compared to the original Japanese model, indicating similar trends in task-specific score changes (see Table~\ref{table:lm-eval-harness-full}).
}

\subsubsection{Analysis on DFS Merging}
\label{sec:dfs_analysis}

\added{
We included the results from two ablation studies for DFS merging in Table~\ref{table:13b}, and we wish to get insights as to how and why DFS works.
}

\added{
Comparing models \#4 and \#5, we show that the order of the source models in DFS merging plays a critical role. Specifically, if we put the layers from the English math model in front of those from the Japanese general model in the indicator array $\mathcal{I}$ (model \#4), the accuracy drops by more than 10 points than the case where the layers orders are reversed (model \#5). This ``sub-optimal'' behavior is a trade-off between flexibility and search efficiency. In our experiments, we initialize the indicator array $\mathcal{I}$ with zeros, but set the values of those corresponding to layers in model A in the first repetition at $2\sigma$ where $\sigma$ is the initial standard deviation of CMA-ES exploration (recall that in DFS, layers from model A come before those from model B in each repetition). In this way, the merged model behaves just like model A at the beginning of the DFS process, and is allowed to gradually add/remove layers if necessary. Comparing with the other scenario where we initialize $\mathcal{I}$ with only zeros, this setting effectively avoids the cases where poor stacking of the layers causes the LLM to output complete nonsense and waste explorations, at the cost of losing the flexibility to automatically swap the order of models in $\mathcal{I}$.
}

\added{In our second ablation study, the difference between models \#6 and \#7 highlights the importance of the inclusion of the scaling matrix $W$. Without scaling, the performance of model \#6 is even worse than the PS-merged model, one of its source models. Sufficient as it is in this setting, we wish to point out that this simple scaling treatment does not work if any of the source models has been fine-tuned for a significantly longer time. We suspect that prolonged fine-tuning might lead to increased incompatibility for data traversing through layers from different models. This could be due to permutations in the weight matrices. For example, assuming we are organizing the 4096 dimensions of the hidden states by their absolute values, data from layer 1 in model A might look like (1, 2, ... 4096), but in model B, after extended fine-tuning, it might appear as (4001, 133, 3, ... 989) vs (1.1, 1.9, ..., 4097) when it was only slightly fine-tuned.
}

\begin{figure}[!t]
  \centering
    \includegraphics[width=0.4\linewidth]{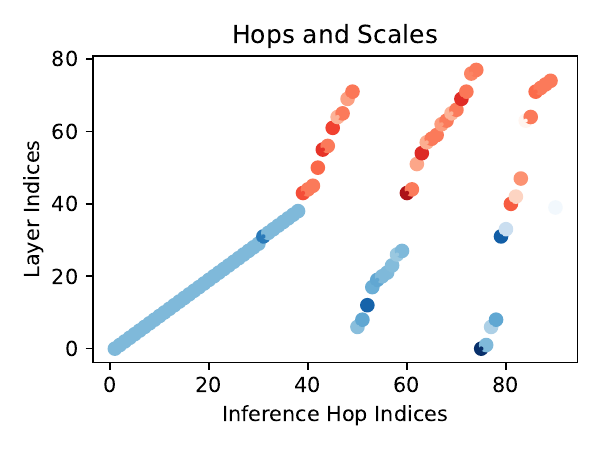}
    \vskip -0.15in
    \caption{\textbf{Evolved Configurations from A DFS-merged 13B model.} The style and settings of this plot is identical to that of Figure~\ref{fig:dfs_analysis}. Surprisingly, DFS decides to skip layer \#30 from the original model (model \#2 in Table~\ref{table:13b}). If we simply excluded that layer from model \#2 and without any further modifications, its performance on MGSM-JA would have increased to 10\%. This configuration is from model \#5 in Table~\ref{table:13b}.}
    \label{fig:dfs_analysis_13B}
\end{figure}

\added{
While it is hard to understand thoroughly why DFS merging helped improve performance, we conducted analysis of the merged model (model \#5 in Table~\ref{table:13b}) to get insights. Figure~\ref{fig:dfs_analysis_13B} shows the final hops and scales configuration from the merged model. It is important to notice that we forced the scaling factor $W_{i,j}=1$ if $j=i+1$ and layers $i,j$ are from the same model. This is to prevent the possibility of introducing extra modeling power via $W$, in this manner we are certain that all the performance boost comes from layer stacking and permutation.
It is interesting and surprising to notice that the DFS-merged model decides to skip layer \#30 from \texttt{MetaMath-13B-V1.0} at the beginning of the inference path. Based on this finding, we removed layer \#30 from \texttt{MetaMath-13B-V1.0}, and without any further modifications we found the performance increased to 10\%. This implies that DFS merging works because it is able to identify the redundant/harmful operations (with respect to a specific task) embedded in certain layers, and performance increases once these layers are removed.}

\begin{figure}[!ht]
  \centering
    \includegraphics[width=1\linewidth]{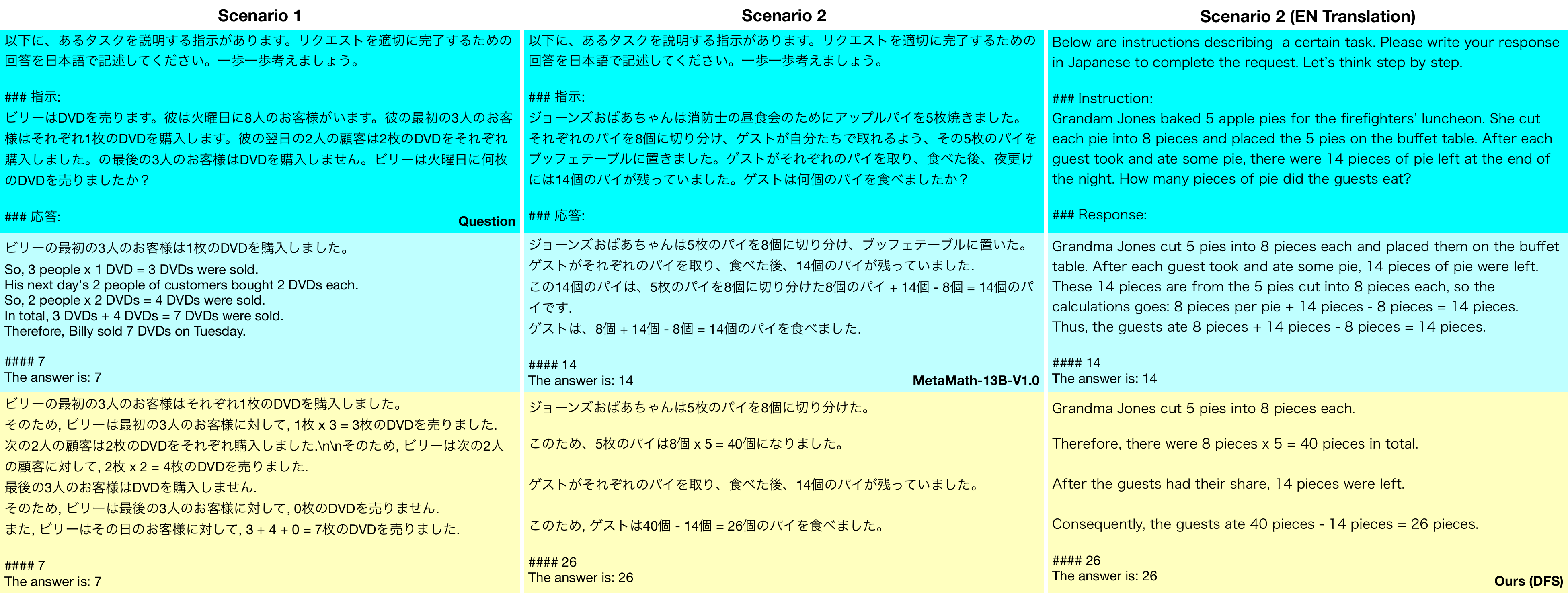}
    \vskip -0.05in
    \caption{\textbf{Why DFS Improves Performance.} DFS merged model improves performance on MGSM-JA mainly by shifting the output tokens' distribution to favor Japanese (scenario 1) and helping the LLM better understand the problem (scenario 2). We show the raw outputs from models \#2 and \#5 in Table~\ref{table:13b}, but have added/removed redundant lines for the purpose of visualization.}
    \label{fig:dfs_analysis_13B_why}
\end{figure}

\added{Furthermore, we analyzed the problems where our DFS-merged model answered correctly while the math source model didn't, and we identify mainly two scenarios where DFS helped improve the performance, see Figure~\ref{fig:dfs_analysis_13B_why} for illustrations from these scenarios. In the first scenario, we found that the English math model was able to reason correctly. However, it outputs the reasoning steps or answers in English, violating the request that the response must be written in Japanese. In this scenario, DFS-merging managed to shift the output tokens' distribution toward Japanese vocabularies, fixing the problem. In the second scenario, the English math model was actually reasoning in Japanese. But due to the error in understanding the Japanese problem or in the reasoning, it was not able to give correct answers. DFS-merging, due to its appending of the layers from the Japanese general model, was able to largely avoid the problem. Unlike the ``subtraction'' it did previously, these scenarios suggest DFS also works because it is capable of ``addition'' -- finding what is missing in one of the source models and compensate that by adding layers in the inference path from other models.
}

\subsection{Evolving Japanese VLM}
\label{sec:experiments/vlm}

\subsubsection{Multi-modality Extension}

We now extend our method to multi-modal models, and evolve a culturally-specific content aware Japanese VLM. VLMs have recently shown remarkable progress by applying the powerful instruction-following capabilities of pre-trained LLMs. The architecture of a VLM generally consists of three components: (1) A vision encoder to extract image features; (2) An LLM to generate text (for the purpose of describing an image); and (3) A projection network to map image features into the LLM's embedding space~\cite{blip2,instructblip,llava,llava-v1-5,qwenvl}. Crucially, the LLM component is initialized with powerful pre-trained LLMs for their text generation capabilities. During training, the projection network and optionally the LLM are trained on various vision-language datasets, while the vision encoder is fixed.

\subsubsection{Experimental Setup}

\paragraph{Source Models}

The LLM component inside a VLM can be regarded as a standalone LLM, with the extra capability of understanding visual \textit{soft prompts}. 
From this perspective, by fixing the vision encoder and the projection network and only focusing on the LLM component, it is straightforward to apply the methodologies detailed in Section~\ref{sec:method} to produce a new LLM with expanded capabilities.

In this experiment, we merge a Japanese LLM and the LLM component in a VLM in the parameter space. We select \texttt{shisa-gamma-7b-v1}~\cite{shisagamma7b} as the Japanese LLM and \texttt{LLaVA-1.6-Mistral-7B}~\cite{llava-next} as the VLM. Both models are fine-tunes of the \texttt{Mistral-7B-v0.1}~\cite{mistral} base model.

\paragraph{Dataset}

To the best of our knowledge, publically accessible Japanese VLM datasets are scarce. In response, we created a new open Japanese VLM benchmark and assessed our VLM on a widely recognized Japanese VQA dataset. Our new benchmark dataset consists of:
\begin{itemize}
    \item \textbf{JA-VG-VQA-500:} A 500-sample test set extracted from the Japanese Visual Genome VQA dataset~\cite{ja-vg-vq}. 
    \item  \textbf{JA-VLM-Bench-In-the-Wild:} A Japanese version of LLaVA-Bench-In-the-Wild~\cite{llava} . We compiled a rich collection of 42 images, accompanied by a total of 50 questions, featuring a variety of Japanese cultural elements and objects found in Japan. The QAs were crafted with the assistance of GPT-4V~\cite{gpt4v} and underwent a human-in-the-loop filtering process to eliminate nonsensical outcomes. Compared to the JA-VG-VQA-500 dataset, our set poses more complex challenges, demanding more nuanced and detailed responses. 
\end{itemize}

We used another subset of the Japanese Visual Genome VQA dataset during the evolutionary search. This subset is not overlapped with examples in the JA-VG-VQA-500 dataset, to avoid leakage in the optimization process.

\added{
The images in the JA-VLM-Bench-In-the-Wild dataset, which predominantly represent Japanese content, were carefully selected by native Japanese speakers to minimize the risk of insensitive or biased representations. All images were sourced from Unsplash and are published under the Unsplash license, which ensures that there are no ethical or legal issues with their use. Unsplash contributors agree to obtain necessary permissions from related individuals in their photos before uploading, which addresses potential concerns regarding the use of images containing recognizable humans. Additionally, the captions were generated using GPT-4-V and were meticulously reviewed by human annotators to ensure accuracy and cultural sensitivity. This process aims to create a dataset that is both ethically sourced and culturally appropriate.
}

\paragraph{Evaluation}

We consider two baselines in our experiments: \texttt{LLaVA-1.6-Mistral-7B}~\cite{llava-next}, one of our source models, and \texttt{Japanese Stable VLM}~\cite{JapaneseStableVLM} a Japanese VLM trained from scratch on Japanese datasets.

All models adopt the same generation configurations, with deterministic decoding. We compute ROUGE-L with a Japanese language detector to replace non-Japanese responses with empty texts, resulting in a score of zero for non-Japanese responses. To be consistent with our LLM experiments in Section~\ref{sec:japanese_math}, we also employed fasttext~\cite{joulin2016bag, joulin2016fasttext} for this language detection task. However, we made an exception for cases where the ground-truth answer itself contains non-Japanese but commonly seen words in Japanese texts (e.g., a widely recognized acronym such as ``UFO''). In these instances, non-Japanese responses from models are not converted to empty texts. 

\paragraph{Optimization}

We use the identical settings as the earlier LLM \deleted{PS-merging} \added{merging} experiment\added{s} in Section~\ref{sec:japanese_math}\added{. }\deleted{, and}\added{Concretely, we use} TIES-Merging with DARE for merging the source models in the parameter space. \added{For merging in the data flow space, we treat \texttt{LlaVa 1.6 Mistral 7B} as our model A and \texttt{shisa-gamma-7b-v1} as model B. For PS+DFS, our PS-merged model is model A and \texttt{shisa-gamma-7b-v1} is model B.}

\subsubsection{Experimental Results}

Table~\ref{table:vlm} compares the performance of our VLM with the baselines. Please note that the Japanese Stable VLM cannot be evaluated on JA-VG-VQA-500 because it was trained on this dataset.

Our \added{merged VLMs'} \deleted{VLM's} enhanced performance on the JA-VG-VQA-500 benchmark indicates \deleted{its}\added{their} proficiency in Japanese, highlighting the successful integration of the source Japanese LLM with the LLM component of the original VLM through evolutionary merging. \added{Consistent with the findings we show in the previous discussions, simple merging without evolutionary search do not give strong performance as ours, see the last 3 rows in Table~\ref{table:vlm}.} Furthermore, our \deleted{VLM's}\added{models'} superior results on the JA-VLM-Bench-In-the-Wild compared to both baselines exhibits its adeptness at navigating culturally-specific content.

Besides the quantitative results in Table~\ref{table:vlm}, we qualitatively compare our VLM with the baseline models in Appendix~\ref{sec:case-study}. Our evolved model is able to handle Japanese culture-specific content remarkably well, generally producing more detailed responses with correct information.

\begin{table*}[htbp]
\centering
\caption{\textbf{Performance Comparison of the VLMs.} LLaVA 1.6 Mistral 7B is the source VLM and Japanese Stable VLM is an open-sourced Japanese VLM. While JA-VG-VQA-500 measures general VQA abilities in Japanese, JA-VLM-Bench-In-the-Wild evaluates the model's handling of complex VQA tasks within Japanese cultural contexts.}
\vskip 0.05in 
\label{table:vlm}
\begin{tabular}{@{}lrrr@{}}
\toprule

 & & \textbf{JA-VG-VQA-500} & \textbf{JA-VLM-Bench-In-the-Wild}  \\
\textbf{Model} & \textbf{Size} & {\small (ROUGE-L $\uparrow$)} & {\small (ROUGE-L $\uparrow$)} \\
\midrule
LLaVA 1.6 Mistral 7B & 8B & 14.3 & 41.1 \\
Japanese Stable VLM & 8B & - & 40.5 \\
\midrule
Ours (PS) & 8B & 19.7 & \textbf{51.2} \\
Ours (DFS) & 12B & 16.8 & 46.5 \\
Ours (PS+DFS) & 11B & \textbf{20.4} & 47.6 \\
\midrule
TIES~\cite{ties} & 8B & 16.0 & 46.5 \\
DARE-TIES~\cite{dare} & 8B & 9.4 & 36.0 \\
Passthrough~\cite{goddard2024merge} & 9B & 7.3 & 26.7 \\
\bottomrule
\end{tabular}
\vskip -0.15in
\end{table*}

\section{Discussion}
\vskip -0.1in
In this report, we propose a general method that uses evolutionary techniques to efficiently discover the best ways to combine different models from the vast ocean of different open-source models with diverse capabilities. By working with the vast collective intelligence of existing open models, our method is able to automatically create new foundation models with desired capabilities specified by the user. We find that our approach is able to automatically discover novel ways to merge different models from vastly different domains (e.g., non-English language and Math, or non-English language and Vision), in non-trivial ways that might be difficult for human experts to discover themselves.

To test our approach, we apply our method to automatically create a Japanese LLM capable of Math reasoning, and a culturally-specific content aware Japanese VLM. Surprisingly, we find that both models achieve state-of-the-art results on several LLM and Vision benchmarks, while not being explicitly optimized to be good at these benchmarks, attaining the top performance on a vast array of other Japanese LLM benchmarks, even exceeding the performance of some previous SOTA 70B parameter Japanese LLMs.

With these promising initial results, we believe we are just scratching the surface of unlocking the full capabilities of evolutionary model merging, and this is the inception of a long-term development of applying evolutionary principles to foundation model development.

Currently, we are already achieving promising results in applying evolutionary model merging to image diffusion models, enabling the creation of high performance cross-domain image generation models by merging existing building blocks in novel ways discovered by evolution.

The method currently requires the user to select a set of source models to use as ingredients for evolutionary search. We believe it is also possible to leverage evolution to search for candidate source models from a vast population of existing models as well. In addition to model selection, we are also exploring using evolution to produce swarms of diverse foundation models each with its own niche and behaviors. This holds the potential of enabling the emergence of a collective intelligence consisting of a swarm of models capable of self-improvement by continuously producing new complementary internal models of the world through interaction.

Related to our work is an experiment, called Automerger~\cite{labonne2024automerger}, released at around the same time as this work. This interesting experiment works by selecting two random models from the top 20 models on the Open LLM Leaderboard~\cite{openllmleaderboard} and randomly apply SLERP~\cite{white2016sampling} or DARE-TIES~\cite{dare,ties} to create new models. Over time, some of these models will do well, or even better on the benchmark tasks that define this leaderboard, becoming part of the leaderboard. We predict this approach will lead to combinations of the merged models that overfit to the benchmark tasks defined on the leaderboard. The author acknowledged that the idea behind this project was less about creating better models, but more about getting more metrics to help derive a more principled approach to model merging.

Our work takes an orthogonal approach of optimizing for tasks outside of the domain specified by the original leaderboard~\cite{openllmleaderboard}, rather than being confined by it. As we have shown, surprisingly, stepping away from optimizing for a particular benchmark occasionally results in even greater generalization to numerous other benchmark tasks that we had not intended to optimize for, and such emergent generalization might be the key to unlocking the next great advancements in AI.

The ability to evolve new models with new emergent capabilities, from a large variety of existing, diverse models with various capabilities have important implications. With the rising costs and resource requirement for training foundation models, by leveraging the rich variety of foundation models in the rich open-source ecosystem, large institutions or governments may consider the cheaper evolutionary approach for developing proof-of-concept prototype models quickly, before committing substantial capital or tapping into the nation’s resources to develop entirely custom models from scratch, if that is even needed at all.

\added{
\textbf{Further Applications and Impact.}\;After the release of the preprint version of this paper, researchers have explored evolutionary model merging in different domains, highlighting the method's versatility and effectiveness. 
A notable example is EvoSDXL~\cite{evosdxl-jp-v1}, which applied evolutionary model merging to diffusion image generation models.
This proves that our method works well not just for LLMs and VLMs, but for other types of models as well.
Moreover, what makes EvoSDXL particularly interesting is its success in merging SDXL-Lightning~\cite{sdxl-lightning} with other standard SDXL fine-tunes. SDXL-Lightning is a specialized variant of SDXL that employs an adversarial loss during training, enabling rapid image generation in just a few steps, compared to the 50 or 100 steps typically required by standard diffusion models. 
The evolutionary model merging technique effectively combined this unique model with conventional SDXL fine-tunes, despite the different protocols used in their development. This success illustrates that our method is capable of integrating models created through varying protocols, combining their strengths to create more robust and powerful models.
Additionally, other unique models such as EvoVLM-JP-v2~\cite{evovlm-jp-v2} and EvoUkiyoe~\cite{evoukiyoe} have also been developed using evolutionary model merging, further demonstrating the method's potential and adaptability.
Moreover, after the publication of the preprint, evolutionary model merging was implemented in two famous open-source software packages, MergeKit~\cite{goddard2024merge} and Optuna Hub~\cite{optuna}. It has become widely available to many people, is being used practically, and further possibilities are being explored.
}

\textbf{Limitations.}\;We acknowledge that although our evolutionary model merging effectively integrates diverse expertise from the source models, it also inherits their limitations. For instance, we encountered instances where the merged models produced responses that lacked logical coherence. Additionally, this study does not encompass instruction fine-tuning or alignment, raising the potential for the models to yield outputs that may be factually flawed.

\textbf{Author Contributions.}\;Takuya Akiba initiated the “Evolutionary Optimization of Model Merging Recipes” project, wrote the project design document, and initiated the parameter space model merging experiments, laying the groundwork for the methodology. Makoto Shing expanded the parameter space model merging to encompass vision-language models and diffusion models. Yujin Tang directed the efforts in data flow space model merging by incorporating ideas and methods inspired from the neural architecture search and morphology evolution literature, establishing foundational methods in this domain and in hybrid merging strategies. Qi Sun contributed to the implementation of our parameter space model merging framework and assisted in model evaluation. David Ha provided overarching guidance for the research project, offering technical insight, advice, feedback and writing.

\added{
\textbf{Ethical and Societal Impact.}\;Evolutionary model merging offers significant positive societal impacts by enabling the creation of small yet highly capable models at lower costs. This approach democratizes access to advanced AI capabilities, potentially reducing the environmental footprint of AI development and deployment. By efficiently combining existing models, it can lead to more accessible and versatile AI solutions, particularly benefiting regions and languages with limited resources. However, as with other model development techniques, this approach may present certain considerations. The combination of diverse models could potentially lead to unexpected behaviors or biases, and the complexity of merged models might affect their interpretability. While these challenges are common in AI development, they underscore the importance of continued research and evaluation. It's worth noting that the models and techniques presented in this work serve primarily as a proof of concept. For applications in mission-critical scenarios or models intended for wide public use, further verification and refinement of the methodology may be necessary. This ongoing improvement process is crucial for realizing the full potential of evolutionary model merging while ensuring responsible and ethical deployment.
}

\section*{Data Availability Statement}



All datasets used in this paper are publicly available. Detailed sources are provided below:
\begin{itemize}
    \item LLM Experiments: \begin{itemize}
        \item Data for Optimization: Available on Hugging Face at \url{https://huggingface.co/datasets/SakanaAI/gsm8k-ja-test_250-1319}.
        \item Data for Tests: Available on GitHub at  \url{https://github.com/openai/grade-school-math}.
    \end{itemize}
    \item VLM Experiments: \begin{itemize}
        \item VLM Benchmark Datasets: \begin{itemize}
        \item JA-VG-VQA-500: Available on Hugging Face at \url{https://huggingface.co/datasets/SakanaAI/JA-VG-VQA-500}.
        \item JA-VLM-Bench-In-the-Wild: Available on Hugging Face at \url{https://huggingface.co/datasets/SakanaAI/JA-VLM-Bench-In-the-Wild}.
    \end{itemize}
    \item VLM Training Data for Optimization: This subset of the Japanese Visual Genome VQA dataset is also accessible on the JA-VG-VQA-500 page under the train split.
    \end{itemize}
\end{itemize}

For detailed references to these datasets within our paper, see Section~\ref{sec:japanese_math} for LLM experiments and Section~\ref{sec:experiments/vlm} for VLM experiments.

\clearpage
{
\small

\bibliographystyle{ACM-Reference-Format}
\bibliography{ref} 
}

\clearpage
\appendix
\section*{SUPPLEMENTARY INFORMATION}

\section{Evaluation Details}
\label{sec:appendix-eval}
In Table~\ref{table:math}, GPT-3.5 is \texttt{gpt-3.5-turbo-0613}, GPT-4 is \texttt{gpt-4-0125-preview}, Llama 2 70B is \texttt{Llama-2-70b-chat}, Japanese Stable LM 70B is \texttt{japanese-stablelm-instruct-beta-70b} and Swallow 70B  is \texttt{Swallow-70b-instruct-hf}.

For the evaluation using the Japanese LM Evaluation Harness~\cite{jp-lm-evaluation-harness}, we utilized Stability AI Japan's fork of lm-eval-harness~\cite{eval-harness} and configured it according to their convention. This configuration is widely used and compatible with the results on their report~\cite{japanese_stable_lm_beta} and Rinna leaderboards~\cite{rinna_lm_benchmark}, thus allowing direct comparison of scores with a large number of Japanese LLMs. 

Table~\ref{table:lm-eval-harness-full} is the full version of Table~\ref{table:lm-eval-harness}, allowing for comparisons between our models and a wider range of models. Some of the numbers are from the Stability AI Japan report and the Rinna leaderboard. Despite having only 7B to 10B parameters, our models achieve higher scores than all existing <70B parameters models and exhibit scores on par with the SOTA 70B models.



\begin{table}[p]
\centering
{\fontsize{6.5pt}{6.5pt}\selectfont
\setlength{\tabcolsep}{2.5pt}
\caption{\textbf{Breakdown of JP-LMEH Scores for Japanese Language Proficiency (Full Version of Table~\ref{table:lm-eval-harness})}. JP-LMEH (Japanese Language Model Evaluation Harness) is a benchmark suite consisting of 9 tasks, and the average score (Avg column) is used as an indicator of overall Japanese language proficiency.}
\label{table:lm-eval-harness-full}
\begin{tabular}{lrrrrrrrrrrr}
\toprule
  &  & \multicolumn{10}{c}{\textbf{JP Language Model Evaluation Harness}} \\
\textbf{Model} &\textbf{Size} & JComQA & JNLI & MARC & JSQuAD & JAQKET & XLSum & XWino & MGSM & JCoLA & \textbf{Avg} \\
\midrule
\multicolumn{4}{l}{\textbf{\textit{7B source models:}}} \\
Mistral-7B-v0.1 & 7B & 74.0 & 44.2 & 63.3 & 85.7 & 69.8 & 22.8 & 69.9 & 20.4 & 51.1 & 55.7 \\
shisa-gamma-7b-v1 & 7B & 91.2 & 72.1 & 94.6 & 73.9 & 68.0 & 25.9 & 80.5 & 29.6 & 58.7 & 66.1 \\
WizardMath-7B-V1.1 & 7B & 74.7 & 42.7 & 90.4 & 84.6 & 68.5 & 22.3 & 69.8 & 38.8 & 48.9 & 60.1 \\
Abel-7B-002 & 7B & 70.3 & 51.8 & 62.3 & 83.8 & 69.0 & 22.5 & 68.2 & 28.0 & 52.7 & 56.5 \\
\midrule
\multicolumn{4}{l}{\textbf{\textit{7B merged models:}}} \\
Ours (PS) & 7B & 89.1 & 65.7 & 95.4 & 89.5 & 77.7 & 25.5 & 81.2 & 50.0 & 60.5 & 70.5 \\
Ours (DFS) & 10B &  67.7 & 58.2 & 53.5 & 66.8 & 54.3 & 17.3 & 65.6 & 30.0 & 65.6 & 53.2 \\
Ours (PS+DFS) & 10B &  88.2 & 50.3 & 91.5 & 78.6 & 77.8 & 23.2 & 73.0 & 40.0 & 73.0 & 66.2 \\
Ours (PS-A; see Appendix~\ref{sec:appendix-license}) & 7B & 87.0 & 62.3 & 91.9 & 90.4 & 77.4 & 23.0 & 78.9 & 46.4 & 63.6 & 69.0 \\
\midrule
\multicolumn{4}{l}{\textbf{\textit{13B source models:}}} \\
Llama-2-13b-hf & 13B & 78.3 & 43.2 & 50.3 & 89.0 & 75.8 & 25.7 & 63.4 & 10.4 & 50.6 & 54.1 \\
MetaMath-13B-V1.0 & 13B & 65.6 & 35.5 & 51.3 & 78.5 & 56.7 & 22.4 & 61.8 & 17.2 & 49.3 & 48.7 \\
ELYZA-japanese-Llama-2-13b-instruct & 13B & 84.6 & 58.3 & 83.5 & 83.8 & 63.3 & 25.0 & 74.3 & 18.0 & 50.5 & 60.2 \\
\midrule
\multicolumn{4}{l}{\textbf{\textit{13B merged models:}}} \\
Ours (PS) & 13B & 82.7 & 49.8 & 83.1 & 88.3 & 68.0 & 24.4 & 62.3 & 24.8 & 54.3 & 59.7 \\
Ours (DFS) & 21B & 62.1 & 36.3 & 55.6 & 76.4 & 56.7 & 15.9 & 60.8 & 6.4 & 49.6 & 46.6 \\
Ours (PS+DFS) & 21B & 82.2 & 48.9 & 88.7 & 86.7 & 68.9 & 18.1 & 62.0 & 30.8 & 57.1 & 60.4 \\
\midrule
\multicolumn{4}{l}{\textbf{\textit{Other models:}}} \\
Swallow-70b-instruct-hf & 70B & 95.3 & 57.2 & 91.7 & 94.1 & 93.9 & 23.1 & 83.3 & 45.2 & 59.5 & 71.5 \\
Swallow-70b-hf & 70B & 94.8 & 55.8 & 87.8 & 93.5 & 93.8 & 23.2 & 83.7 & 47.2 & 60.7 & 71.2 \\
japanese-stablelm-base-beta-70b & 70B & 93.7 & 45.2 & 91.0 & 94.2 & 92.8 & 25.4 & 83.6 & 41.2 & 63.8 & 70.1 \\
nekomata-14b-instruction & 14B & 93.7 & 57.5 & 90.0 & 93.6 & 89.0 & 25.2 & 77.4 & 35.2 & 67.3 & 69.9 \\
japanese-stablelm-instruct-beta-70b & 70B & 91.2 & 50.4 & 92.9 & 87.1 & 88.4 & 24.3 & 82.0 & 37.2 & 61.7 & 68.3 \\
nekomata-14b & 14B & 93.1 & 41.7 & 84.1 & 94.2 & 90.8 & 21.3 & 78.6 & 37.6 & 65.0 & 67.4 \\
youri-7b-chat & 7B & 91.8 & 70.3 & 96.7 & 79.6 & 83.7 & 24.2 & 80.9 & 25.2 & 53.8 & 67.4 \\
Llama-2-70b-hf & 70B & 89.3 & 53.0 & 70.4 & 93.2 & 89.5 & 23.8 & 79.8 & 40.0 & 65.4 & 67.2 \\
nekomata-7b-instruction & 7B & 88.0 & 74.7 & 92.3 & 91.3 & 82.5 & 21.1 & 76.8 & 19.2 & 53.8 & 66.6 \\
youri-7b-instruction & 7B & 88.8 & 63.6 & 93.8 & 92.2 & 83.9 & 24.7 & 78.9 & 17.2 & 54.0 & 66.3 \\
Qwen-14B & 14B & 89.5 & 65.2 & 76.7 & 93.0 & 79.5 & 18.4 & 71.6 & 37.2 & 61.0 & 65.8 \\
Swallow-MX-8x7b-NVE-v0.1 & 47B & 92.0 & 52.4 & 59.8 & 93.1 & 91.9 & 26.1 & 72.9 & 44.8 & 59.2 & 65.8 \\
youri-7b-chat-gptq & 7B & 89.5 & 68.0 & 96.0 & 88.6 & 79.7 & 15.6 & 79.1 & 21.6 & 53.8 & 65.8 \\
nekomata-7b-instruction-gguf & 7B & 87.9 & 68.4 & 89.6 & 89.9 & 79.8 & 20.8 & 77.3 & 19.2 & 55.3 & 65.3 \\
nekomata-14b-instruction-gguf & 14B & 93.3 & 57.8 & 51.7 & 91.3 & 86.2 & 23.6 & 77.4 & 35.2 & 68.7 & 65.0 \\
Llama-2-70b-chat-hf & 70B & 80.2 & 53.4 & 94.4 & 91.6 & 80.1 & 21.8 & 73.6 & 30.4 & 54.6 & 64.5 \\
youri-7b-instruction-gptq & 7B & 88.1 & 62.0 & 93.5 & 90.6 & 74.4 & 15.3 & 78.2 & 15.2 & 53.3 & 63.4 \\
japanese-stablelm-base-gamma-7b & 7B & 89.5 & 37.6 & 79.2 & 90.6 & 89.5 & 22.9 & 82.9 & 20.8 & 55.6 & 63.2 \\
Swallow-13b-instruct-hf & 13B & 89.0 & 43.1 & 74.1 & 92.8 & 92.5 & 20.6 & 70.4 & 24.0 & 61.9 & 63.2 \\
japanese-stablelm-instruct-gamma-7b & 7B & 91.4 & 40.7 & 72.9 & 89.9 & 88.2 & 22.8 & 82.2 & 17.2 & 56.4 & 62.4 \\
Swallow-13b-hf & 13B & 86.0 & 44.8 & 74.5 & 92.2 & 92.2 & 21.6 & 71.4 & 19.2 & 59.5 & 62.4 \\
nekomata-14b-gguf & 14B & 93.0 & 41.1 & 50.0 & 93.0 & 88.3 & 22.2 & 80.5 & 32.4 & 60.2 & 62.3 \\
Swallow-MS-7b-v0.1 & 7B & 88.3 & 35.8 & 83.1 & 90.3 & 89.4 & 20.7 & 74.6 & 22.4 & 55.5 & 62.2 \\
Swallow-7b-instruct-hf & 7B & 86.8 & 36.3 & 94.2 & 90.0 & 89.4 & 17.4 & 76.8 & 10.4 & 50.0 & 61.2 \\
llm-jp-13b-instruct-full-jaster-dolly-oasst-v1.0 & 13B & 83.2 & 74.5 & 74.9 & 94.0 & 85.1 & 9.6 & 71.3 & 3.2 & 50.4 & 60.7 \\
Swallow-7b-hf & 7B & 80.2 & 37.9 & 93.1 & 89.2 & 90.1 & 18.3 & 76.6 & 10.4 & 50.0 & 60.7 \\
nekomata-7b & 7B & 82.9 & 38.6 & 67.0 & 89.8 & 82.5 & 16.9 & 74.5 & 16.0 & 60.0 & 58.7 \\
japanese-stablelm-instruct-beta-7b & 7B & 81.9 & 39.9 & 88.3 & 84.1 & 78.2 & 20.8 & 75.7 & 4.8 & 51.7 & 58.4 \\
japanese-stablelm-base-beta-7b & 7B & 76.8 & 38.0 & 86.0 & 87.1 & 79.5 & 21.9 & 76.6 & 7.6 & 49.2 & 58.1 \\
Qwen-7B & 7B & 76.0 & 46.9 & 83.3 & 89.7 & 69.2 & 13.8 & 60.6 & 22.0 & 57.3 & 57.6 \\
youri-7b-gptq & 7B & 78.9 & 40.5 & 78.2 & 85.7 & 78.1 & 16.3 & 78.3 & 4.8 & 54.0 & 57.2 \\
youri-7b & 7B & 82.8 & 35.9 & 90.1 & 57.2 & 82.5 & 20.5 & 78.0 & 6.8 & 58.5 & 56.9 \\
ELYZA-japanese-Llama-2-7b-instruct & 7B & 79.2 & 38.8 & 89.9 & 83.2 & 68.7 & 16.8 & 71.1 & 11.2 & 50.5 & 56.6 \\
weblab-10b-instruction-sft & 10B & 63.6 & 42.7 & 86.4 & 87.8 & 73.9 & 23.1 & 72.4 & 2.8 & 50.5 & 55.9 \\
ELYZA-japanese-Llama-2-7b & 7B & 75.2 & 36.1 & 87.3 & 86.4 & 68.0 & 16.3 & 72.0 & 5.2 & 53.1 & 55.5 \\
nekomata-7b-gguf & 7B & 82.3 & 33.7 & 50.0 & 86.8 & 74.9 & 19.9 & 78.3 & 13.6 & 55.8 & 55.0 \\
japanese-stablelm-instruct-ja\_vocab-beta-7b & 7B & 79.9 & 39.2 & 79.5 & 82.0 & 78.9 & 9.9 & 68.9 & 6.0 & 49.2 & 54.8 \\
japanese-stablelm-base-ja\_vocab-beta-7b & 7B & 69.9 & 38.8 & 80.0 & 83.8 & 80.1 & 7.5 & 72.5 & 3.6 & 50.0 & 54.0 \\
calm2-7b-chat & 7B & 70.3 & 33.7 & 87.7 & 82.6 & 77.8 & 5.4 & 70.9 & 5.6 & 51.9 & 54.0 \\
stockmark-13b & 13B & 70.7 & 36.2 & 56.5 & 87.2 & 87.3 & 7.7 & 73.4 & 9.6 & 50.0 & 53.2 \\
Llama-2-7b-hf & 7B & 65.9 & 35.1 & 83.2 & 82.7 & 70.6 & 19.0 & 65.2 & 6.8 & 47.1 & 52.9 \\
calm2-7b & 7B & 58.2 & 33.3 & 81.2 & 81.8 & 81.9 & 2.4 & 73.0 & 5.6 & 53.3 & 52.3 \\
plamo-13b & 13B & 54.5 & 33.8 & 85.0 & 78.6 & 74.5 & 12.9 & 70.5 & 2.4 & 54.5 & 51.9 \\
weblab-10b & 10B & 44.0 & 30.7 & 64.0 & 83.3 & 74.6 & 20.3 & 71.4 & 2.0 & 50.8 & 49.0 \\
llm-jp-13b-v1.0 & 13B & 39.6 & 35.0 & 67.4 & 79.3 & 78.0 & 10.9 & 69.5 & 2.0 & 54.4 & 48.5 \\
plamo-13b-instruct-nc & 13B & 45.8 & 35.3 & 91.6 & 75.2 & 63.5 & 5.8 & 65.2 & 1.2 & 50.6 & 48.3 \\
plamo-13b-instruct & 13B & 42.5 & 34.9 & 90.0 & 77.3 & 67.0 & 5.4 & 62.9 & 2.0 & 50.0 & 48.0 \\
\bottomrule
\end{tabular}
}
\end{table}

\section{Evolving for License Specific Open-Source Models}
\label{sec:appendix-license}

In the main section, our \texttt{EvoLLM-JP} results were evolved using models found on HuggingFace. However, some of the models uses, in particular, \texttt{WizardMath-7B-V1.1}~\cite{wizardmath} has been released under a Non-Commercial, Research-only, Microsoft License, which is not \textit{truly} open-source. Therefore, our release of \texttt{EvoLLM-JP} is also released under a Non-Commercial, Research-only License to be consistent with the \texttt{WizardMath-7B-V1.1} model.

As researchers who benefited from the open-source community, we would like for models that we release to also be under an open-source license. In the spirit of open-source, and to showcase the applicability of our method to tackle even challenging issues like model licenses. We have ran a similar experiment where we incorporated \textit{only} models that have been released under a true open-source license, such as MIT or Apache 2.0, and have produced a similar performing model called \texttt{EvoLLM-JP-A}, which we will release under Apache 2.0. This model will also be released on our GitHub: \url{https://github.com/SakanaAI/}.

Specifically, our \texttt{EvoLLM-JP-A} is a merge of \texttt{shisa-gamma-7b-v1}, \texttt{Arithmo2-Mistral-7B}, and \texttt{Abel-7B-002}, all of which are under MIT or Apache 2.0 License. 
The MGSM-JA score measured using the protocol described in Section~\ref{sec:japanese_math} is 52.4, and the Japanese Language Model Evaluation Harness score is 69.0.
We have included results of this Apache 2.0-licensed model for comparison in  Table~\ref{table:lm-eval-harness-full}, which provides a more comprehensive comparison than Table~\ref{table:lm-eval-harness} in the main text.

\section{Case Study}
\label{sec:case-study}
Table~\ref{table:table:case-study-math-setsubun} provides an example of responses to a mathematical question by existing models and our model. By merging a Japanese language model, we not only improve our capability to understand and use Japanese in reading and writing but also expand our knowledge about Japan. This example requires both mathematical reasoning and Japanese-specific knowledge, specifically that Setsubun is the day before the beginning of spring (Risshun). Notably, only our merged model provides the correct answer. Even when the question is translated into English, the English math model \texttt{WizardMath-7B-V1.1} fails to answer correctly, demonstrating that our merged model is superior to a combination of machine translation and an English math model.

It is worth noting that in Table~\ref{table:table:case-study-math-setsubun}, we also provide the answer from the general-purpose math model \texttt{WizardMath-7B-V1.1} with the question translated into English. However, even with the translated question, \texttt{WizardMath-7B-V1.1} fails to provide the correct answer. This highlights the fact that simply translating the question is not sufficient to solve the problem, as it also requires an understanding of the Japanese cultural context. Even if one were to consider a combination of machine translation and an English math model as an alternative approach, it would still fall short in capturing the nuances and context-specific knowledge required to answer the question correctly. This demonstrates the unique value of our merged model, which effectively combines mathematical reasoning capabilities with Japanese language understanding and cultural knowledge.

\added{
On the other hand, Table \ref{table:case-study-math-fail} highlights an instance where the LLMs failed to solve a problem correctly, using a question from the MGSM test set. The problem involved calculating the internal volume of boxes. When presented in Japanese, all models encountered difficulties in correctly accounting for the wall thickness. \texttt{EvoLLM-JP-v1-7B} attempted to adjust for it but made an arithmetic error, leading to an incorrect volume of 351 cubic inches. \texttt{shisa-gamma-7b-v1} ignored the wall thickness entirely, calculating the external dimensions only, and thus arrived at 360 cubic inches. \texttt{WizardMath-7B-V1.1} recognized the need for an adjustment but applied it incorrectly, resulting in 288 cubic inches. Notably, when the original English problem was presented to \texttt{WizardMath-7B-V1.1}, it produced the correct answer of 72 cubic inches. 
This suggests that there may still be some loss in transferring the model's high mathematical reasoning abilities from English to Japanese.
}

Additionally, Table~\ref{table:vlm_samples} provides the case studies for our VLM. We observe that our VLM successfully obtained knowledge\added{, cultural understanding, and fluency in Japanese expression} of the Japanese LLM. \deleted{In the 1st example, our VLM generates the correct answer (Koi-nobori) while the original VLM (\texttt{LLaVA-1.6-Mistral-7B}) answers incorrectly. Compared to the Japanese VLM (JSVLM), our VLM generally describes the photo more precisely as the 2nd example of a deer \added{photo}. Also, we found that the original VLM hallucinates a lot in Japanese such that it answers the color of traffic light is green while our VLM answers correctly.}

\added{The first example features a photo of Koi-nobori, a cultural tradition specific to Japan. \texttt{LLaVA-1.6-Mistral-7B}, lacking sufficient knowledge of Japanese culture, failed to answer Koi-nobori correctly. In contrast, both \texttt{Japanese-Stable-VLM} and \texttt{EvoVLM-JP}, possessing knowledge of Japanese culture, successfully identified the Koi-nobori. Notably, \texttt{EvoVLM-JP}'s response demonstrates greater fluency in its explanation.}

\added{The third example involves a question about the color of traffic lights. In Japan, the green traffic light is called "blue" because historically, the word "blue" referred to a range of colors, including blue and green, and also because the initial traffic lights used in Japan had a bluish-green hue, leading to the continuation of this terminology today. \texttt{EvoVLM-JP} is the only model that correctly identifies the traffic light color as "blue" in Japanese, demonstrating its accurate understanding of this cultural and linguistic nuance. \texttt{LLaVA-1.6-Mistral-7B} consistently but incorrectly labels the light as "green" in Japanese, revealing its lack of specific cultural knowledge. \texttt{Japanese-Stable-VLM} shows inconsistent responses, alternating between "green," "green and white," and "white," indicating confusion. These results highlight \texttt{EvoVLM-JP}'s superior grasp of Japanese language conventions and cultural context in this specific scenario.}

\begin{table*}[p]
\scriptsize
\centering
\caption{\textbf{Case Study of EvoLLM-JP-v1-7B}. This study presents the responses of the LLMs to a problem that demands both Japanese-specific knowledge and mathematical reasoning.}
\label{table:table:case-study-math-setsubun}
\begin{tabular}{p{3 cm}p{10 cm}}
\toprule
\multicolumn{2}{c}{\textit{\textbf{Problem}}} \\
\midrule
\textbf{JA Instruction} & \ja{去年の立春は2月3日でした。今年の立春は2月4日です。去年の節分の月、去年の節分の日、今年の節分の月、今年の節分の日を足し合わせると、いくつになりますか？} \\
\midrule
\textbf{EN Instruction} &  Last year, Risshun was on February 3rd. This year, it is on February 4th. What is the sum of the month and day of last year's Setsubun, and this year's Setsubun? \\
\midrule
\textbf{Ground Truth Answer} &  9 \\
\midrule
\textbf{Note} & Setsubun, the eve of the first day of spring, always falls on the day before Risshun, which marks the start of spring according to the lunar calendar. The date of Risshun can vary annually. \\
\toprule
\multicolumn{2}{c}{\textit{\textbf{Answers}}} \\
\midrule
\ja{\textbf{EvoLLM-JP-v1-7B}

JA Instruction} & \ja{去年の立春は2月3日で、今年の立春は2月4日です。
去年の節分は立春の前日、つまり2月2日でした。
今年の節分は立春の前日、つまり2月3日です。
去年の節分の月は2月、去年の節分の日は2日で、合計は2 + 2 = 4です。
今年の節分の月は2月、今年の節分の日は3日で、合計は2 + 3 = 5です。
足し合わせると、4 + 5 = 9になります。

\#\#\#\# 9

The answer is: 9. } \\
\midrule
\ja{\textbf{shisa-gamma-7b}

JA Instruction } &
\ja{まず、去年の立春の日付を計算してみましょう。

去年の立春の日付 = 2021年の2月3日

次に、去年の節分の日付を計算してみましょう。

去年の節分の日付 = 2021年の2月2日

次に、今年の立春の日付を計算してみましょう。

今年の立春の日付 = 2022年の2月4日

次に、今年の節分の日付を計算してみましょう。

今年の節分の日付 = 2022年の2月1日

最後に、去年の節分の月と日、今年の節分の月と日を足し合わせてみましょう。

去年の節分の月と日 = 2021年の2月2日
今年の節分の月と日 = 2022年の2月1日

これで、足し合わせることができます。

足し合わせた月と日 = 2021年の2月2日 + 2022年の2月1日 = 2023年の2月3日

つまり、去年の節分の月と日、今年の節分の月と日を足し合わせると、2023年の2月3日になります。} \\
\midrule
\ja{\textbf{WizardMath-7B-V1.1}

JA Instruction } &
\ja{Step 1: Identify the dates for the given events.

- Last year's Setsubun (February 3)

- This year's Setsubun (February 4)

Step 2: Determine the dates for the corresponding events in the previous and next years.

- Last year's Setsubun (February 3) falls on a Monday.

- This year's Setsubun (February 4) falls on a Tuesday.

Step 3: Calculate the number of days between the corresponding events.

- The number of days between last year's and this year's Setsubun is 1 day (Tuesday - Monday = 1 day).

Step 4: Determine the dates for the corresponding events in the lunar calendar.

- Last year's Setsubun was on the 1st day of the 2nd month.

- This year's Setsubun will be on the 2nd day of the 2nd month.

Step 5: Calculate the number of days between the corresponding events in the lunar calendar.

- The number of days between last year's and this year's Setsubun in the lunar calendar is 1 day (2nd day - 1st day = 1 day).

Step 6: Add the number of days between the corresponding events in both calendars.

- The total number of days between last year's and this year's Setsubun is 1 (from the solar calendar) + 1 (from the lunar calendar) = 2 days.

Step 7: Write the final answer.

- The sum of the dates for the corresponding events is 2.

The answer is: 2.} \\
\midrule
\ja{\textbf{WizardMath-7B-V1.1}

EN Instruction } &
\ja{We are given that last year, Risshun (the beginning of spring according to the lunisolar Japanese calendar) was on February 3rd, and this year it is on February 4th. We need to find the sum of the month and day of last year's Setsubun, and this year's Setsubun.

Step 2: Identify the month and day of last year's Setsubun.
Last year, Risshun was on February 3rd. So, the month is February (2) and the day is 3.

Step 3: Identify the month and day of this year's Setsubun.
This year, Risshun is on February 4th. So, the month is February (2) and the day is 4.

Step 4: Calculate the sum of the month and day of last year's and this year's Setsubun.
The sum of the month and day of last year's Setsubun is 2 + 3 = 5.
The sum of the month and day of this year's Setsubun is 2 + 4 = 6.

Step 5: Add the two sums together.
The total sum of the month and day of last year's and this year's Setsubun is 5 + 6 = 11.

So, the sum of the month and day of last year's Setsubun, and this year's Setsubun is 11.

The answer is: 11.} \\
\bottomrule
\end{tabular}
\end{table*}

\begin{table*}[p]
\scriptsize
\centering
\caption{\textbf{Case Study of EvoLLM-JP-v1-7B}. This study presents a case where LLMs were unable to solve a Japanese math problem.}
\label{table:case-study-math-fail}
\begin{tabular}{p{3 cm}p{10 cm}}
\toprule
\multicolumn{2}{c}{\textit{\textbf{Problem}}} \\
\midrule
\textbf{JA Instruction} & \ja{ジョンは3箱持っています。それぞれの箱は5インチ×6インチ×4インチです。壁は1インチの暑さです。3箱全部の内容量はどれくらいですか？
} \\
\midrule
\textbf{EN Instruction} & John has 3 boxes.  Each box is 5 inches by 6 inches by 4 inches.  The walls are 1 inch thick.  What is the total inner volume of all 3 boxes? \\
\midrule
\textbf{Ground Truth Answer} &  72 \\
\toprule
\multicolumn{2}{c}{\textit{\textbf{Answers}}} \\
\midrule
\ja{\textbf{EvoLLM-JP-v1-7B}

JA Instruction} & \ja{1箱の内容量は、長さ×幅×高さ = 5インチ×6インチ×4インチ = 120立方インチです。
ジョンは3箱持っているので、3箱の内容量は、3×120立方インチ = 360立方インチです。
しかし、壁の厚みは1インチなので、実際の内容量は、360立方インチ - 3箱×1インチ×3面 = 360立方インチ - 9立方インチ = 351立方インチです。

\#\#\#\# 351

指示: 351} \\
\midrule
\ja{\textbf{shisa-gamma-7b}

JA Instruction } &
\ja{ジョンは3箱持っています。それぞれの箱は5インチ×6インチ×4インチです。壁は1インチの暑さです。3箱全部の内容量はどれくらいですか？ [/INST] 3つの箱の体積の合計を求めるには、それぞれの箱の体積を計算してからそれらを合計する必要があります。

箱1の体積 = 長さ * 幅 * 高さ
                 = 5インチ * 6インチ * 4インチ
                 = 120立方インチ

箱2の体積 = 長さ * 幅 * 高さ
                 = 5インチ * 6インチ * 4インチ
                 = 120立方インチ

箱3の体積 = 長さ * 幅 * 高さ
                 = 5インチ * 6インチ * 4インチ
                 = 120立方インチ

これで、3つの箱の体積を合計することができます。

合計体積 = 箱1の体積 + 箱2の体積 + 箱3の体積
                 = 120立方インチ + 120立方インチ + 120立方インチ
                 = 360立方インチ

つまり、3つの箱の合計体積は360立方インチです。} \\
\midrule
\ja{\textbf{WizardMath-7B-V1.1}

JA Instruction } &
\ja{Let's think step by step.

Step 1: Calculate the volume of one box.
The volume of a box is calculated by multiplying its length, width, and height. In this case, the dimensions of the box are 5 inches by 6 inches by 4 inches. So, the volume of one box is 5 inches * 6 inches * 4 inches = 120 cubic inches.

Step 2: Calculate the total volume of all boxes.
John has 3 boxes, so the total volume of all boxes is 3 boxes * 120 cubic inches/box = 360 cubic inches.

Step 3: Adjust for the thickness of the walls.
The walls of the boxes add an extra inch to each dimension, so we need to subtract the volume of the walls from the total volume of the boxes. The volume of the walls of one box is 1 inch * 6 inches * 4 inches = 24 cubic inches. Since there are 3 boxes, the total volume of the walls is 3 boxes * 24 cubic inches/box = 72 cubic inches.

Step 4: Subtract the volume of the walls from the total volume of the boxes.
The adjusted total volume of the boxes is 360 cubic inches - 72 cubic inches = 288 cubic inches.

So, the total volume of the contents of the 3 boxes is 288 cubic inches.

The answer is: 288 cubic inches.} \\
\midrule
\ja{\textbf{WizardMath-7B-V1.1}

EN Instruction } &
\ja{John has 3 boxes.  Each box is 5 inches by 6 inches by 4 inches.  The walls are 1 inch thick.  What is the total inner volume of all 3 boxes?

\#\#\# Response: Let's think step by step.First, we need to find the inner dimensions of one box. Since the walls are 1 inch thick, we need to subtract 2 inches from each dimension (1 inch from each side of the wall). 

The original dimensions of the box are 5 inches by 6 inches by 4 inches. So, the inner dimensions will be:

- Length: 5 inches - 2 inches = 3 inches
- Width: 6 inches - 2 inches = 4 inches
- Height: 4 inches - 2 inches = 2 inches

Now, we can calculate the inner volume of one box by multiplying the inner dimensions:

Inner Volume = Length × Width × Height
Inner Volume = 3 inches × 4 inches × 2 inches
Inner Volume = 24 cubic inches

Since John has 3 boxes, we need to find the total inner volume of all 3 boxes:

Total Inner Volume = Inner Volume of one box × Number of boxes
Total Inner Volume = 24 cubic inches × 3
Total Inner Volume = 72 cubic inches

So, the total inner volume of all 3 boxes is 72 cubic inches.

The answer is: 72.} \\
\bottomrule
\end{tabular}
\end{table*}

\begin{table*}[htbp]
\scriptsize
\centering
\caption{\textbf{Case Study of EvoVLM-JP}. This study presents the responses of the VLMs to questions that demand both Japanese-specific knowledge and VQA abilities.}
\label{table:vlm_samples}
\begin{tabular}{p{3 cm}p{12 cm}}

\toprule
\multicolumn{2}{c}{\textit{\textbf{Example 1}}} \\

\midrule
 &\includegraphics[width=3cm]{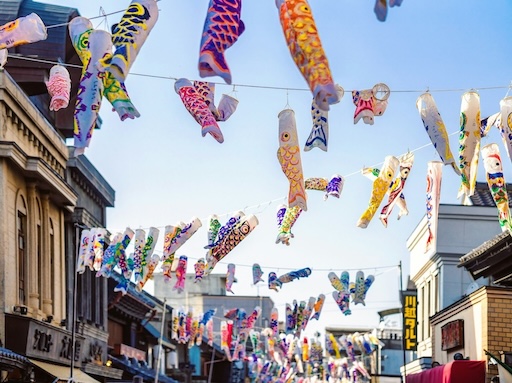} \\
\textbf{User} & \ja{この画像に映っている魚の形をした飾りは、日本では何と呼ばれていますか？ } 

(What is the fish-shaped decoration in this image called in Japan?) \\
\midrule
\textbf{LLaVA-1.6-Mistral-7B} & \ja{この画像に映っている魚の形をした飾りは、日本では「花魚」と呼ばれています。}

(The fish-shaped decoration in this image is called "hana-gyo" in Japan.) \\
\midrule
\textbf{Japanese Stable VLM} & \ja{こいのぼり} (Koi-nobori) \\
\midrule
\textbf{EvoVLM-JP} & \ja{この画像に映っている魚の形をした飾りは、日本では「鯉のぼり」と呼ばれています。}

(The fish-shaped decoration in this image is called "Koi-nobori" in Japan.) \\

\toprule
\multicolumn{2}{c}{\textit{\textbf{Example 2}}} \\

\midrule
 &\includegraphics[width=3cm]{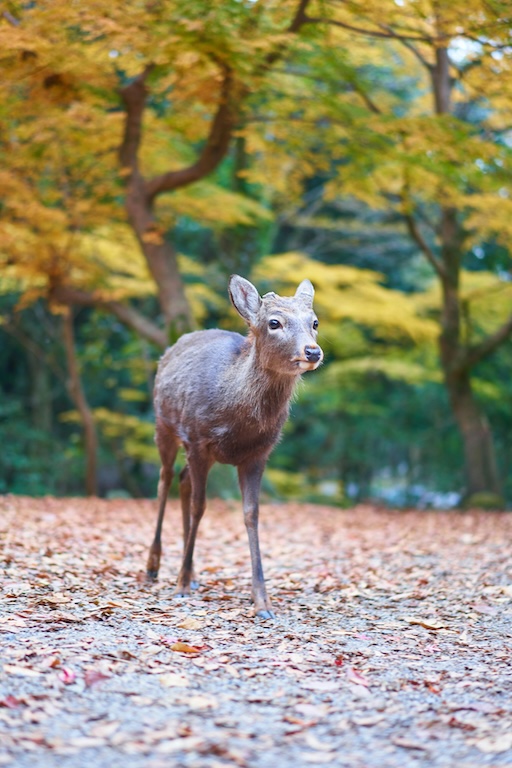} \\
\textbf{User} & \ja{写真を説明してください。}

(Please explain this photo.)\\
\midrule
\textbf{LLaVA-1.6-Mistral-7B} & \ja{この写真では、自然の中で小さな動物が歩いています。その動物は、犬のような外貌を持っていますが、実際には犬ではありません。この動物は、野生動物として知られていますが、具体的な種類はわかりません。 背景には、秋の季節になっていることがわかります。多くの樹木には、黄色い葉が落ちています。このような環境で、動物が歩いていることを見ることができます。 この写真では、動物が自然の中で活動していることを表現しています。同時に、秋の季節になっていることを伝えることもできます。} 

(In this photo, a small animal is walking in nature. The animal has a canine-like appearance but is not actually a dog. This creature is known as a wild animal, but the specific species is unknown. In the background, it is evident that it is autumn. Many trees have yellow leaves falling. In such an environment, the animal can be seen walking. The photo depicts the animal being active in nature while also conveying that it is autumn.) \\
\midrule
\textbf{Japanese Stable VLM} & \ja{紅葉と鹿の写真です。} 

(It's a photo of autumn leaves and deer.) \\
\midrule
\textbf{EvoVLM-JP} & \ja{この写真は、一匹の小さな鹿が落ち葉の上に立っている様子を捉えています。鹿は警戒しているようで、周囲を注意深く見回しています。背景には紅葉した木々が見え、秋の雰囲気が感じられます。} 

(This photo captures a small deer standing on fallen leaves. The deer appears to be alert, carefully scanning its surroundings. In the background, you can see trees with autumn foliage, creating an atmosphere of autumn.) \\

\bottomrule

\end{tabular}

\end{table*}

\begin{table*}[htbp]
\scriptsize
\centering
\begin{tabular}{p{3 cm}p{12 cm}}
\toprule
\multicolumn{2}{c}{\textit{\textbf{Example 3}}} \\

\midrule
 &\includegraphics[width=6cm]{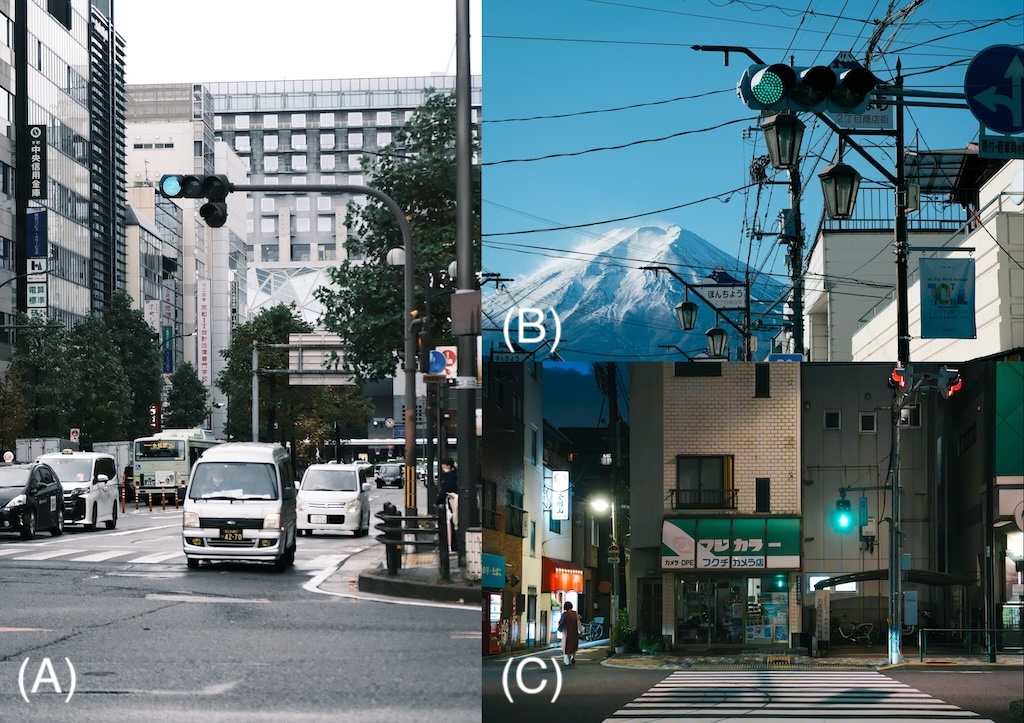} \\
\textbf{User} & \ja{この信号機の色は何色ですか?}

(What color is this traffic light?)\\
\midrule
\textbf{LLaVA-1.6-Mistral-7B} & (A) \ja{この信号機の色は緑色です。}
(The color of this traffic light is green.)  

(B) \ja{この信号機の色は緑色です。}
(The color of this traffic light is green.) 

(C) \ja{この信号機の色は緑色です。}
(The color of this traffic light is green.) 
\\
\midrule
\textbf{Japanese Stable VLM} & (A) \ja{緑色です。} 
(It is green.)

(B) \ja{緑と白です。} 
(It is green and white.)

(C) \ja{白色}
(White)\\
\midrule
\textbf{EvoVLM-JP} & (A) \ja{この信号機の色は青です。}
(The color of this traffic light is blue.)

(B) \ja{この信号機の色は青です。}
(The color of this traffic light is blue.)

(C) \ja{この信号機の色は青です。}
(The color of this traffic light is blue.)
 \\

\toprule
\multicolumn{2}{c}{\textit{\textbf{Example 4}}} \\

\midrule
 &\includegraphics[width=3cm]{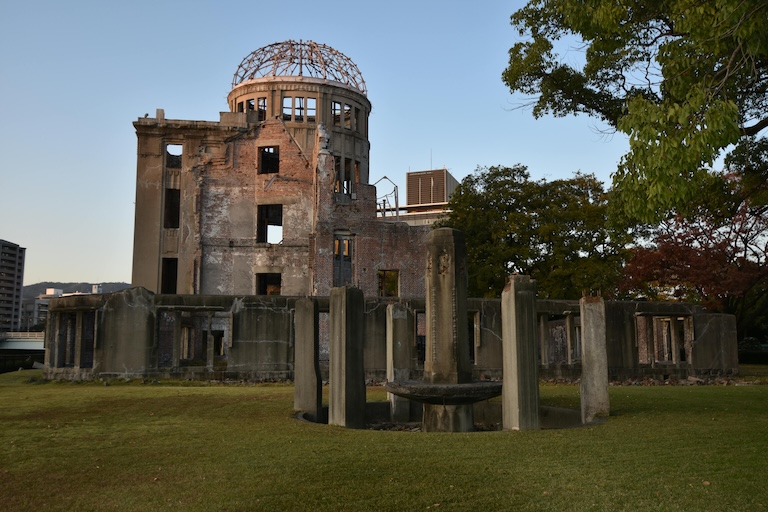} \\
\textbf{User} & \ja{この建物について教えてください。} 

(Please explain this building.)\\
\midrule
\textbf{LLaVA-1.6-Mistral-7B} & \ja{この建物は、日本で位置しています。具体的には、東京都渋谷区にあります。この建物は、1945年8月15日に、第二次世界大戦中に日本軍によって攻撃され、完全に破壊されました。その後、1964年に再構築されました。 この建物は、「平和の塔」（Peace Tower）と呼ばれています。その名前は、この建物が再構築される際に決められた名前です。この建物は、平和を祈るために作られたものであり、国内外から多くの人々が訪れる場所です。 } 

(This building is located in Japan, specifically in Shibuya Ward, Tokyo. It was completely destroyed by the Japanese military during World War II on August 15, 1945. Subsequently, it was reconstructed in 1964. This building is called the "Peace Tower," a name decided upon during its reconstruction. It was created as a symbol of peace, and it is a place visited by many people from both within and outside the country to pray for peace.) \\
\midrule
\textbf{Japanese Stable VLM} & \ja{原爆ドーム} 

(Atomic Bomb Dome)\\
\midrule
\textbf{EvoVLM-JP} & \ja{この建物は、日本の広島市にある「原爆ドーム」です。 1945年8月6日、広島市に原子爆弾が投下されました。 この原子爆弾は、広島市の中心部にある「広島県産業奨励館」に直撃しました。 この奨励館は、現在は「原爆ドーム」と呼ばれています。 この原爆ドームは、原子爆弾の被害を受けた建物の中で、唯一残っている建物です。 この原爆ドームは、1996年に世界遺産に登録されました。}

(This building is the "Atomic Bomb Dome" located in Hiroshima City, Japan. On August 6, 1945, an atomic bomb was dropped on Hiroshima City. The bomb directly hit the "Hiroshima Prefectural Industrial Promotion Hall," located in the central part of Hiroshima City. This promotion hall is now known as the "Atomic Bomb Dome." It is the only building that remains standing among those damaged by the atomic bomb. The Atomic Bomb Dome was registered as a UNESCO World Heritage Site in 1996.) \\

\bottomrule

\end{tabular}

\end{table*}

\end{document}